\documentclass[10pt,twocolumn,letterpaper]{article}

\usepackage[pagenumbers]{cvpr} 

\usepackage{xcolor}

\definecolor{darkgreen}{rgb}{0.0, 0.5, 0.0}
\definecolor{olive}{rgb}{0.4, 0.4, 0.0}
\definecolor{darkred}{rgb}{0.8, 0.0, 0}

\definecolor{cvprblue}{rgb}{0.21,0.49,0.74}
\usepackage[pagebackref,breaklinks,colorlinks,allcolors=cvprblue]{hyperref}

\def\paperID{} 
\def\confName{CVPR}
\def\confYear{2026}

\usepackage{graphicx}
\usepackage{booktabs}

\usepackage[accsupp]{axessibility}  

\usepackage{hyperref}

\usepackage{orcidlink}
\usepackage{multirow}
\usepackage{makecell}
\usepackage{amssymb}
\usepackage{xcolor} 
\usepackage{bbm}
\usepackage{algorithm}
\usepackage{algorithmic}
\usepackage{float}
\newcommand{\scright}[1]{\hfill{#1}}

\begin{document}

\title{One Adapter for All: Towards Unified Representation in Step-Imbalanced Class-Incremental Learning}

\author{Xiaoyan Zhang\\
University of Michigan\\
Ann Arbor, Michigan, U.S.A\\
{\tt\small xyzaxis@umich.edu}
\and
Jiangpeng He\\
Indiana University\\
Bloomington, Indiana, U.S.A\\
{\tt\small jhe2@iu.edu}
}

\maketitle

\begin{abstract}
Class-incremental learning (CIL) aims to acquire new classes over time while retaining prior knowledge, yet most setups and methods assume balanced task streams. In practice, the number of classes per task often varies significantly. We refer to this as step imbalance, where large tasks that contain more classes dominate learning and small tasks inject unstable updates. Existing CIL methods assume balanced tasks and therefore treat all tasks uniformly, producing imbalanced updates that degrade overall learning performance. To address this challenge, we propose One-A, a unified and imbalance-aware framework that incrementally merges task updates into a single adapter, maintaining constant inference cost. One-A performs asymmetric subspace alignment to preserve dominant subspaces learned from large tasks while constraining low-information updates within them. An information-adaptive weighting balances the contribution between base and new adapters, and a directional gating mechanism selectively fuses updates along each singular direction, maintaining stability in head directions and plasticity in tail ones.
Across multiple benchmarks and step-imbalanced streams, One-A achieves competitive accuracy with significantly low inference overhead, showing that a single, asymmetrically fused adapter can remain both adaptive to dynamic task sizes and efficient at deployment. Code is available at \url{https://github.com/xiaoyanzhang1/One-A}.
\end{abstract}

\section{Introduction}
\label{sec:intro}

\begin{figure*}[t]
    \centering
    \includegraphics[width=\linewidth]{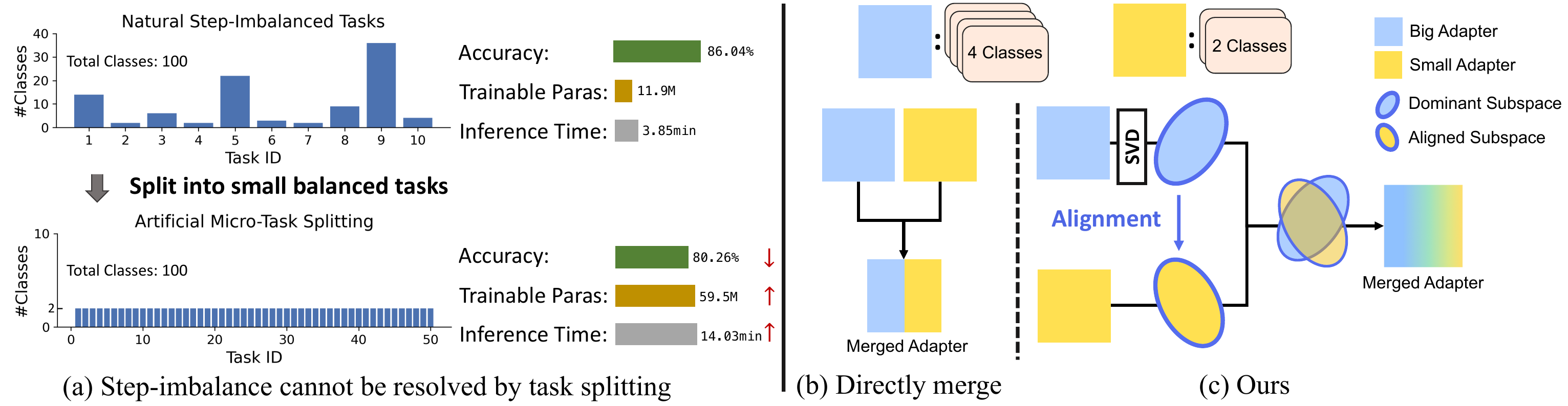}
    \caption{Necessity of step-imbalanced continual learning and overview of our method.
(a) Task splitting increases incremental steps, resulting in higher computation, more parameters, and stronger interference, which harms accuracy. (b) Direct merging averages task-specific parameters without alignment. (c) Our method aligns task subspaces via SVD and performs direction-aware fusion to preserve dominant knowledge while integrating information into a unified adapter with balanced stability and plasticity.} 
    \label{fig:intro}
\end{figure*}

Class-incremental learning (CIL) aims to continuously learn from a stream of tasks while retaining knowledge of previously encountered classes~\cite{li2017learning,castro2018end,rebuffi2017icarl,yoon2017lifelong,dhar2019learning}. The central challenge of CIL lies in overcoming catastrophic forgetting, where updating the model on new tasks often leads to performance degradation on earlier ones\cite{hu2018overcoming,Zhou_2024}. To address these challenges, recent works leverage pre-trained models as backbones and apply lightweight adaptation modules such as prompt tuning~\cite{wang2022learning,wang2022dualprompt,smith2023coda} or adapter tuning~\cite{zhou2025revisiting,zhou2024expandable,liang2024inflora,he2025cl} to inject task-specific knowledge while freezing most parameters, which significantly improves the performance compared to full finetuning\cite{zhou2024continual,kirkpatrick2017overcoming,li2017learning,wang2022beef}.
However, the existing CIL is widely studied under a balanced task scenario where each task contains the same number of classes, which is overly idealized. In real-world applications, new tasks always vary in size. For example, in a clothing store recognition system, seasonal updates may add many new categories at once, whereas daily arrivals contribute only a few. We refer to this practical setting as Step-Imbalanced Class-Incremental Learning (SI-CIL), where a \emph{step} denotes a single incremental update stage and different steps introduce highly varying numbers of new classes. 
A naive way to handle step imbalance is to artificially split large tasks into multiple balanced micro-tasks.
However, as illustrated in Figure~\ref{fig:intro}a, empirical results of EASE~\cite{zhou2024expandable} on CIFAR100~\cite{krizhevsky2009learning} show that this strategy increases the number of incremental steps, leading to higher computational cost and stronger task interference, which ultimately degrades performance. 
Therefore, step imbalance cannot be effectively mitigated by task rebalancing and should be addressed directly.
Moreover, it reflects realistic deployment conditions where the class distribution across tasks is neither uniform nor predictable.

Unlike traditional CIL, SI-CIL introduces significant task-size disparity across tasks. Large tasks provide abundant supervision and stable gradients, while small tasks generate noisy updates that interfere with previously learned knowledge. When all tasks are optimized with equal importance, these noisy updates exacerbate catastrophic forgetting and cause instability.
Most existing CIL methods do not account for such task-size disparity. Adapter-based approaches, although efficient and widely adopted in PTM-based CIL, still optimize each task uniformly. As a result, large tasks dominate the shared representation space, and small tasks contribute marginal, unstable updates, further aggravating the stability–plasticity imbalance. For example, EASE~\cite{zhou2024expandable} and CL-LoRA~\cite{he2025cl} rely on task-specific adapters to ensure stability, yet the poorly learned representations from small tasks can interfere with well-formed knowledge from large tasks. In addition, maintaining a separate adapter for each task increases inference cost and parameter overhead as tasks accumulate. 
While the most recent work~\cite{fukuda2025adapter} improves scalability by merging adapters, its coarse, weight-based fusion ignores the internal structure of each adapter, preventing it from selectively preserving important parameters or adaptively incorporating task-specific information.

To tackle these limitations, we propose an adaptive and efficient model merging framework for step imbalance as illustrated in Figure~\ref{fig:intro}c. Instead of maintaining separate modules for each task, we keep a single adapter throughout the incremental process and incrementally merge task-specific updates. To handle task heterogeneity, we introduce a direction-aware asymmetric merging strategy that preserves the dominant subspace of large tasks while flexibly injecting information from small ones, different from existing directly averaging \cite{fukuda2025adapter} as seen in Figure~\ref{fig:intro}b. This design makes our method both adaptive to dynamic task sizes and efficient in inference. Our contributions are summarized as follows:
\vspace{-2mm}
\begin{itemize}
    \item  We introduce and analyze step-imbalance in class-incremental learning, revealing its unique challenges compared to balanced settings.
    \item We propose a direction-aware asymmetric merging strategy that allocates representational capacity according to task informativeness, enabling effective integration of large and small tasks with a single adapter.
    \item Through extensive experiments on multiple benchmarks and imbalance settings, we demonstrate that our method achieves strong and competitive performance while maintaining high inference efficiency.
\end{itemize}

\section{Related Works}
\label{sec:related_works}

\subsection{Class-Incremental Learning}
\label{subsec:cil}
Class-Incremental Learning (CIL) aims to continually acquire new classes while retaining prior knowledge~\cite{Zhou_2024,9915459,wang2024comprehensive}.  
Traditional CIL methods can be broadly categorized into three groups.  
\emph{Rehearsal-based} approaches~\cite{NEURIPS2019_e562cd9c,iscen2020memory,liu2020mnemonics,rebuffi2017icarl,NIPS2017_0efbe980,luo2023class,wang2022foster} store exemplars from previous classes to mitigate forgetting.  
\emph{Regularization-based} methods~\cite{aljundi2018memory,aljundi2019task,li2017learning} impose stability-promoting penalties to prevent destructive updates.  
\emph{Expansion-based} strategies~\cite{chen2023dynamic,yan2021dynamically,hu2023dense,douillard2020podnet} grow network capacity to encode new information.  
Despite these efforts, learning new classes from scratch while preserving past knowledge remains challenging.

\begin{figure}[t]
    \centering
    \includegraphics[width=1.0\linewidth]{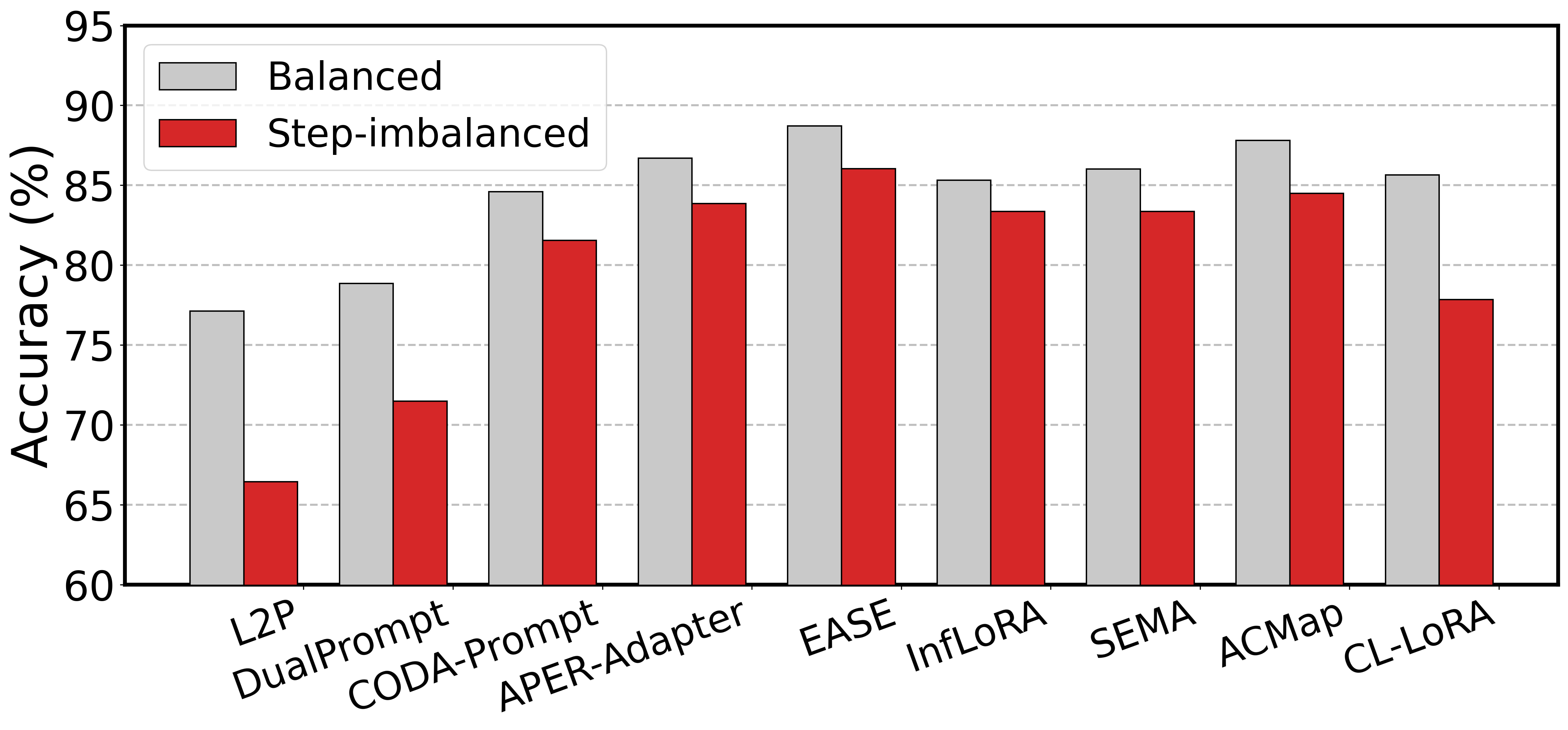}
    \vspace{-0.2cm}
    \caption{Performance comparison under balanced and step-imbalanced settings. All methods show clear degradation under step-imbalanced tasks.}
    \label{fig:balance_imb}
    \vspace{-0.3cm}
\end{figure}

Leveraging powerful pre-trained models (PTMs) such as ViTs~\cite{dosovitskiy2020image} and CLIP~\cite{radford2021learning} has become a popular choice for strong performance and efficient training~\cite{zhou2024continual}.
\emph{Prompt-based CIL}~\cite{wang2022learning,wang2022dualprompt,smith2023coda,wang2022s,jung2023generating} adapts PTMs through learnable prompts. However, when task sizes are step-imbalanced, prompt evolution becomes biased because small early tasks easily overfit and shift the shared prompts, which harms performance as shown in Figure~\ref{fig:balance_imb}.
\emph{Adapter-based CIL} inserts lightweight trainable modules such as LoRA~\cite{zhou2025revisiting,he2025cl,gao2023unified,liang2024inflora} or MLP adapters~\cite{zhou2025revisiting,zhou2024expandable,wang2025self} into ViT layers while keeping backbone frozen. APER-Adapter~\cite{zhou2025revisiting} trains the adapter only on the first task and then freezes it, but its performance heavily depends on the order of tasks, and very small initial tasks can severely hinder learning. EASE~\cite{zhou2024expandable} allocates one adapter for each task but does not account for task-size differences, leading to biased representations. CL-LoRA~\cite{he2025cl} combines shared and task-specific adapters, yet the bias introduced by small tasks gradually accumulates in the shared adapter and influences the learning of later tasks. Other adapter methods~\cite{liang2024inflora,wang2025self} show similar issues under step-imbalanced regimes, where task-size disparity and order effects significantly degrade performance.

\vspace{-3mm}
\paragraph{\textbf{Data-Imbalanced Class-Incremental Learning.}}
Existing works~\cite{liu2022longtailedclassincrementallearning,wang2024long,raghavan2024delta,he2025longtailedcontinuallearningvisual} have explored sample-imbalanced continual learning, where each incremental step introduces a fixed number of classes but the sample distribution across classes follows a long-tailed pattern.
These methods primarily address \emph{sample-level imbalance} by correcting intra-task bias through strategies such as reweighting
\cite{he2024gradient,park2021influence,ren2020balanced,cao2019learning}, re-sampling~\cite{park2022majority,van2007experimental}, or feature decoupling and alignment~\cite{kang2019decoupling,wang2024long}, to retain old knowledge.
In contrast, we study a step-imbalanced setting, where the number of classes per task varies across steps. 
To our knowledge, this form of imbalance has not been systematically formalized or analyzed in prior CIL literature.
By fixing the sample size per class, SI-CIL provides a controlled formulation that disentangles task-level imbalance from sample-level effects, enabling focused analysis of a previously overlooked challenge.

\subsection{Model Merging in Class-Incremental Learning}
\label{subsec:model_fusion}

Recent efforts in CIL have explored model fusion as a parameter-efficient alternative to retraining. ACMap~\cite{fukuda2025adapter} introduces adapter merging where task-specific adapters are trained sequentially and iteratively averaged into a unified adapter. 
While simple and scalable, this averaging-based approach lacks explicit subspace alignment and treats all tasks equally, making it less reliable under step-imbalanced conditions.
MOS~\cite{sun2025mos} adopts an EMA-based adapter merging strategy during training to mitigate parameter drift, together with a self-refined retrieval mechanism during inference to select the most suitable adapter.
Although effective in alleviating forgetting, its iterative retrieval process introduces considerable inference latency, limiting scalability in long task sequences.

Beyond CIL, recent work on model merging explores gradient-free strategies for consolidating task-specific models into a shared representation space. KnOTS~\cite{stoica2024model} exemplifies this paradigm by aligning task-update subspaces via SVD and constructing a common low-rank subspace that captures shared knowledge across models. However, its symmetric design assumes comparable update magnitudes, causing small-task directions to undesirably influence the shared representation when task sizes vary. Building on this perspective, we propose an asymmetric subspace alignment strategy that explicitly accounts for task-size disparity during fusion. This approach preserves dominant directions from large tasks while efficiently injecting new knowledge from small ones, enabling faster, more stable, and adaptive adapter fusion under step-imbalanced settings.

\section{Preliminaries}
\label{sec:preliminaries}
In this section, we introduce the step-imbalanced class-incremental learning scenario and describe the backbone architecture based on MLP adapters and prototype classifier.


\subsection{Step-Imbalanced Class-Incremental Learning}
\label{subsec:sicil}
We consider a class-incremental learning scenario where the number of classes per task is \emph{unequal}, resulting in a step-imbalanced task distribution. Given a sequence of $T$ training sets $\{\mathcal{D}_1, \ldots, \mathcal{D}_T\}$, the $t$-th task $\mathcal{D}_t = \{(x_i, y_i)\}_{i=1}^{N_t}$ consists of input images $x_i \in \mathbb{R}^D$ and labels $y_i \in \mathcal{Y}_t$, where $\mathcal{Y}_t \cap \mathcal{Y}_{t'} = \emptyset$ for $t \neq t'$. 
Unlike conventional CIL, the number of classes $|\mathcal{Y}_t|$ varies across tasks, while the number of samples per class remains fixed. To generate the step-imbalanced distribution, we first compute class ratios $r_k = \gamma^{\frac{k}{C-1}}$ for $k = 0, \ldots, C-1$, where $C$ is the total number of classes and $\gamma$ is the imbalance factor.
Given a total of $C=100$ classes divided into $T=10$ tasks, we compute exponential class ratios $r_k=\gamma^{\frac{k}{C-1}}$ with step imbalance factor $\gamma=0.01$.
These ratios are normalized and used to allocate different numbers of classes to each task from head to tail, so that head tasks contain over $35$ classes, while tail tasks include only $1$–$3$.
To emulate realistic non-sequential arrivals, we then randomly permute the task order, producing the final step-imbalanced sequence. A visualization of the generation procedure can be found in Section~\ref{sec:more_sicil} in the Appendix.

At incremental step $t$, we target the rehearsal-free setup~\cite{zhou2025revisiting,he2025cl}: only $\mathcal{D}_t$ is available for training, while the accumulated label space is $\mathcal{Y}_{1:t} = \mathcal{Y}_1 \cup \cdots \cup \mathcal{Y}_t$. The objective is to learn a unified classifier $f(x): \mathcal{X} \to \mathcal{Y}_{1:t}$ minimizing the expected error over all seen classes,
\[
f^* = \arg\min_{f \in \mathcal{H}} \; \mathbb{E}_{(x,y)\sim \mathcal{D}_{1:t}}\big[\mathbbm{1}(y \ne f(x))\big],
\]
where $\mathcal{H}$ is the hypothesis space and $\mathbbm{1}(\cdot)$ is the indicator function.
Following typical PTM-based CIL settings~\cite{wang2022learning,wang2022dualprompt,zhou2024expandable}, a pre-trained ViT~\cite{dosovitskiy2020image} with a linear classification head is often used as the backbone. In our work, we replace this linear head with a prototype-based classifier which is consistent with \cite{he2024gradient,zhou2025revisiting}, and incorporate lightweight task adapters into the backbone.

\subsection{Backbone with MLP Adapters and Prototype Classifier}
\label{subsec:backbone}

In rehearsal-free CIL, we adopt a pre-trained vision transformer (ViT)~\cite{dosovitskiy2020image} as the backbone feature extractor $f_\theta(\cdot)$~\cite{wang2022learning,zhou2025revisiting,he2025cl}. To enable parameter-efficient adaptation, lightweight MLP adapters $f_{\text{adapt}}(\cdot)$ are inserted into the FFN blocks while keeping all pre-trained weights frozen. Each adapter consists of a down-projection $W_{\mathrm{down}} \in \mathbb{R}^{d \times r}$, a non-linear activation (ReLU), and an up-projection $W_{\mathrm{up}} \in \mathbb{R}^{r \times d}$. 
For an input image $x$, the frozen backbone produces a hidden feature $h = f_\theta(x)$, 
and the adapter-augmented residual output is 
\begin{equation}
z = h + \mathrm{ReLU}(h W_{\mathrm{down}})\, W_{\mathrm{up}}.
\end{equation}
For each new task $t$, we instantiate a fresh set of adapters $\{ f_{\text{adapt}}^{(t)} \}$, 
while previously trained adapters remain frozen.

Following prior work~\cite{zhou2025revisiting,he2025cl}, we adopt a prototype-based classifier instead of a unified linear head at inference. During training of task $t$ with class set $\mathcal{C}_t$, we optimize a local linear head $w^{(t)}_\phi \in \mathbb{R}^{d \times |\mathcal{C}_t|}$ on the current data to provide a discriminative signal for the adapters:
\begin{equation}
\min_{f_{\text{adapt}}^{(t)},\, w^{(t)}_\phi}
\frac{1}{|\mathcal{D}_t|} \sum_{(x,y)\in\mathcal{D}_t}
\mathcal{L}_{\mathrm{CE}}\!\big(w^{(t)}_\phi(z),\, y\big),
\end{equation}
where $z$ is the adapter-augmented feature. After training, the local head is discarded and we compute class prototypes
\begin{equation}
\mathcal{D}_t^c := \{(x,y)\in\mathcal{D}_t:\, y=c\}, \quad
p_c^{(t)} \;=\; \frac{1}{|\mathcal{D}_t^c|} \sum_{(x,y)\in \mathcal{D}_t^c} z.
\end{equation}
At inference, we use cosine similarity with $\ell_2$-normalized features and prototypes:
\begin{equation}
\hat{y} \;=\; \arg\max_{\,c \in \bigcup_{i=1}^{t} \mathcal{C}_i}
\big\langle \hat{p}_c^{(i)},\, \hat{z}^{(i)}(x) \big\rangle,
\qquad
\end{equation}

In our method, task-specific adapters are subsequently merged into a \emph{single} adapter via asymmetric, scale-adaptive fusion, so inference only uses one fused adapter while retaining prototype-based classification.

\section{Methodology}
\label{sec:method}

\begin{figure*}[t]
    \centering
    \includegraphics[width=\linewidth]{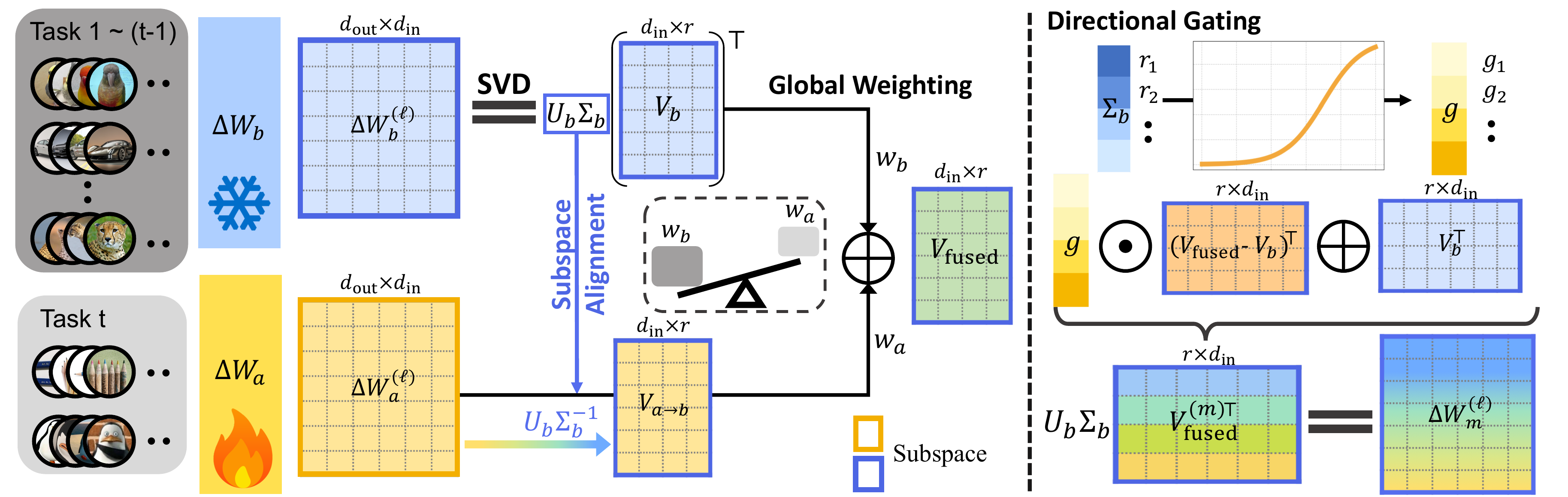}
    \caption{Overview of One-A. For each layer, we decompose the large-task adapter via SVD to extract its dominant subspace $U_b\Sigma_b$ and project the small-task adapter for subspace alignment ($V_{a\rightarrow b}$). A global fusion is first performed between aligned components to obtain $V_{\mathrm{fused}}$, followed by direction-wise gating $g_i$ on the update $(V_{\mathrm{fused}} - V_b)$, which adaptively modulates each singular direction to balance stability and plasticity and yields the final merged adapter $\Delta W_{m}$.}
    \label{fig:framework}
\end{figure*}

Our method aims to address the subspace misalignment and representation bias introduced by step-imbalanced class-incremental learning, where large and small tasks coexist. The overall procedure is summarized in Algorithm~\ref{alg:onea}. As illustrated in Figure~\ref{fig:framework}, our framework incorporates an \emph{asymmetric subspace alignment mechanism} together with two \emph{scale-adaptive fusion strategies} to achieve stable subspace alignment and balanced information integration. Specifically, we build on the fundamental MLP adapter framework described in Section~\ref{subsec:backbone}. In the post-processing stage of each task $t$, we first perform asymmetric SVD to define the dominant subspace based on large-task updates and constrain smaller tasks to update only within this subspace. Next, information-adaptive weighting computes task-level information scores to dynamically balance the influence of old and new tasks during model fusion. Subsequently, directional gating performs fine-grained fusion along each singular direction. It preserves critical information from major subspaces while flexibly injecting new knowledge from minor directions, effectively mitigating catastrophic forgetting. After each task, we merge adapters into a single one, so only one adapter is used at inference time. 
This avoids the growing computational cost typical of multi-adapter systems while still preserving task-relevant knowledge through asymmetric fusion. Algorithmic pseudocode for the One-A merging pipeline is provided in Section~\ref{subsec:code} of Appendix.

\begin{algorithm}[t]
\caption{Continual Learning with One-A}
\label{alg:onea}
\begin{algorithmic}[1]

\REQUIRE Task sequence $\{\mathcal{D}_1,\ldots,\mathcal{D}_T\}$, pretrained backbone $\Theta$
\ENSURE Unified adapter $\Delta W_T$


\FOR{$t = 1$ to $T$}

    \STATE Initialize task adapter $\Delta W_t^{\mathrm{new}}$
    \STATE Train $\Delta W_t^{\mathrm{new}}$ on $\mathcal{D}_t$

    \IF{$t = 1$}
        \STATE $\Delta W_t \leftarrow \Delta W_t^{\mathrm{new}}$
    \ELSE
        \STATE $\Delta W_t \leftarrow \text{OneA\_Merge}(\Delta W_{t-1},\, \Delta W_t^{\mathrm{new}})$
    \ENDIF

    \STATE $\mathcal{P}_t \leftarrow \text{ComputePrototypes}(\Theta + \Delta W_t,\, \mathcal{Y}_{t})$
    \STATE Update classifier $f_t$ using $(\Theta + \Delta W_t)$ and $\mathcal{P}_t$

\ENDFOR

\RETURN $\Delta W_T$

\end{algorithmic}
\end{algorithm}

\subsection{Subspace Alignment under Imbalanced Tasks}
\label{subsec:SAS}
Recent gradient-free fusion methods such as KnOTS~\cite{stoica2024model} mitigate subspace misalignment by aligning task-update before merging.
Given task-specific updates matrics ${\Delta W}$, KnOTS performs singular value decomposition (SVD) on their concatenation to extract a shared basis and merges the aligned components within it.
While effective, this symmetric treatment does not account for task-size disparity, allowing small-task updates to undesirably influence the dominant subspace of large tasks and cause subspace drift.


To handle subspace distortion under step-imbalanced tasks, we adopt an \emph{asymmetric} alignment strategy that freezes the dominant subspace of the larger task and projects the smaller task onto it before fusion. When a new task $t$ arrives, we compare its data volume $|\mathcal{D}_t|$ with the cumulative size of previous tasks $\sum_{i=1}^{t-1} |\mathcal{D}_i|$ and select the larger one as the base adapter $b$ and the smaller one as the align adapter $a$.
For each adapter layer $\ell$, we apply SVD to the base adapter parameter 
$\Delta W_b^{(\ell)} \in \mathbb{R}^{d_{\text{out}}\times d_{\text{in}}}$:
\begin{equation}
\Delta W_b^{(\ell)} = U_b \Sigma_b V_b^\top,
\label{eq:base}
\end{equation}
where $U_b \in \mathbb{R}^{d_{\text{out}}\times r}$, $\Sigma_b \in \mathbb{R}^{r\times r}$, and 
$V_b \in \mathbb{R}^{d_{\text{in}}\times r}$ define the dominant subspace determined by the large task. 
Unlike KnOTS, we freeze $U_b \Sigma_b$ and restrict the smaller task to this space. 
Concretely, the align adapter $\Delta W_a^{(\ell)}$ is projected into the base right-singular space via
\begin{equation}
V_{a \rightarrow b} \;=\; \big((U_b \Sigma_b)^+ \Delta W_a^{(\ell)}\big)^\top 
\;=\; \Delta W_a^{(\ell)\top} U_b\, \Sigma_b^{-1} \;\in\; \mathbb{R}^{d_{\text{in}}\times r},
\label{eq:adapter_align}
\end{equation}
ensuring that the smaller task does not rotate or redefine the dominant subspace.

This asymmetric design ensures that large-task subspaces remain stable, while small tasks contribute only in complementary directions, effectively mitigating subspace drift under step-imbalanced scenarios.

\subsection{Information-Adaptive Global Weighting}
\label{subsec:GW}

While asymmetric subspace alignment prevents small tasks from distorting the dominant subspace, it does not control the global fusion direction. In practice, not all tasks contribute equally, particularly under step-imbalanced scenarios where large and small tasks contain vastly different amounts of information. To address this, we introduce an information-adaptive global weighting strategy that  balances the contribution of the base and align adapters during fusion.

We assign global weights $w_b$ and $w_a$ based on relative information content:
\begin{equation}
w_a = \frac{\phi(\mathrm{Info}_a)}{\phi(\mathrm{Info}_a)+\phi(\mathrm{Info}_b)}, 
\quad
w_b = 1 - w_a,
\label{eq:global_weight}
\end{equation}
where $\phi(\cdot)$ denotes an information proxy. In principle, $\phi$ can reflect various statistics such as singular value energy, representation variance, or class entropy. 

\textbf{Practical choice.} We compare alternative proxies including Frobenius-norm energy and singular-value energy in Section~\ref{subsec:proxy_variants}, and class-count–based weighting achieves the best performance in SI-CIL setting. For stable and reproducible estimation, we adopt the number of classes as the information proxy, which correlates well with task diversity under imbalanced settings and requires no extra computation: 
\begin{equation}
\phi(\mathrm{Info}_t) = \#\mathrm{classes}(t).
\label{eq:weight_class}
\end{equation}

Given these weights, the global fusion of the right singular components is expressed as
\begin{equation}
V_{\mathrm{fused}} = w_b V_b + w_a V_{a \rightarrow b},
\label{eq:global_fusion}
\end{equation}
where $V_b$ denotes the base adapter's singular components and $V_{a \rightarrow b}$ the projected align adapter. This coarse-grained weighting controls whether the fused representation leans more toward the dominant or supplementary task, providing an interpretable and adaptive global balancing mechanism.

\subsection{Directional Gating}
\label{subsec:DG}
Although global weighting can roughly balance the contributions of large and small tasks, applying a single scalar weight to all directions creates a fundamental trade-off: preserving dominant directions and injecting new knowledge cannot be achieved simultaneously. Let $\varepsilon$ denote the tolerated drift on the base dominant subspace, and $\tau$ the required injection ratio of the align task. Global weights $w_b$ and $w_a$ must satisfy
\begin{equation}
(1 - \varepsilon) + \tau \le 1 \quad \Leftrightarrow \quad \tau \le \varepsilon,
\end{equation}
indicating that higher plasticity inevitably reduces stability.

To overcome this limitation, we assign different fusion strengths to different singular directions. Intuitively, high-energy directions should be preserved more conservatively, while low-energy directions can flexibly incorporate new information. Formally, for each singular direction $i$, we compute a gate $g_i \in [0,1]$ that controls how much the align adapter contributes:
\begin{equation}
(V_{\mathrm{fused}}^{(g)})_i = (V_b)_i + g_i \big[(V_{\mathrm{fused}})_i - (V_b)_i\big].
\label{eq:dirc_fuse}
\end{equation}
Here, $g_i = 0$ fully preserves the base direction, while $g_i = 1$ fully adopts the align update.
Based on the SVD decomposition $\Delta W^{(b)}_\ell = U_b \Sigma_b V_b^\top$ introduced above,
let $\Sigma_b = \mathrm{diag}(s_1,\dots,s_r)$, where $s_1$ is the largest singular value.
We determine $g_i$ from the normalized singular values of the base adapter:
\begin{equation}
\tilde{s}_i = \frac{s_i}{s_1 + \delta}, \qquad 
\theta = Q_q(\tilde{S}), \qquad
g_i = \sigma\big(\kappa (\theta - \tilde{s}_i)\big),
\label{eq:directional_weight}
\end{equation}
where $\tilde{S}=\{\tilde{s}_i\}_{i=1}^r$, $Q_q(\cdot)$ denotes the $q$-quantile,
$\delta$ is a small constant for numerical stability, and $\sigma(\cdot)$ is a sigmoid gate. The final merged update is
\begin{equation}
\begin{aligned}
V_{\mathrm{fused}}^{(m)} = V_b + g \odot (V_{\mathrm{fused}} - V_b), \\
\Delta W^{(\mathrm{m})} = U_b \Sigma_b V_\mathrm{fused}^{(m)\top}.
\label{eq:directional_fusion}
\end{aligned}
\end{equation}
This mechanism allows fine-grained control over fusion strength across directions, effectively balancing stability and plasticity without introducing manual thresholds.

\subsection{Optimization Objective}
\label{subsec:optimization}
To improve representation learning for tasks of different sizes, we incorporate a contrastive loss~\cite{1640964} as an auxiliary objective. This loss encourages features from the same class to cluster together while pushing apart features from different classes, providing additional structural constraints.
Since small tasks contain fewer classes and limited supervision, the cross-entropy loss alone may not provide sufficient guidance for stable and discriminative representation learning. To address this, we assign a larger weight to the contrastive loss for smaller tasks, and gradually reduce it as task size increases. This adaptive weighting provides stronger regularization where class information is scarce, leading to more robust representations in step-imbalanced settings. The training objective is
\begin{equation}
\mathcal{L} = \big(1 - \lambda(t)\big)\mathcal{L}_{\mathrm{CE}}(\text{logits}, y)
+ \lambda(t)\mathcal{L}_{\mathrm{ctr}}(F, y),
\label{eq:loss}
\end{equation}
where $\lambda(t)$ is task-dependent, with larger values for smaller tasks. Explicit explanation refers to Section~\ref{subsec:adaptive_training} in the supplementary.

All components work together to achieve stable and scale-aware subspace fusion, enabling effective incremental learning under step-imbalanced scenarios.

\section{Experiments}
\subsection{Implementation Details}

\textbf{Datasets: }
We conduct experiments on widely used benchmarks for class-incremental learning, including CIFAR100~\cite{krizhevsky2009learning}, CUB200~\cite{wah2011caltech}, ImageNet-A~\cite{hendrycks2021natural}, and ImageNet-R~\cite{hendrycks2021many}. 
CIFAR100 contains 100 common object classes, while CUB200 consists of 200 fine-grained bird species. 
ImageNet-A and ImageNet-R each contain 200 classes and exhibit a significant domain gap compared to the original ImageNet~\cite{russakovsky2015imagenet}, which is used for pretraining. 
Specifically, ImageNet-A consists of challenging natural images selected from ImageNet, and ImageNet-R provides artistic renditions of ImageNet images. 
Each dataset is divided according to the number of tasks $T$ and step-imbalance ratio $\gamma$ as described in Section~\ref{subsec:sicil}.
Further details are provided in the Section~\ref{subsec:detail_setting}.

\noindent \textbf{Compared Methods: }
We compare our approach against state-of-the-art, exemplar-free, PTM-based CIL methods, including L2P~\cite{wang2022learning}, DualPrompt~\cite{wang2022dualprompt}, CODA-Prompt~\cite{smith2023coda}, SimpleCIL~\cite{zhou2025revisiting}, APER~\cite{zhou2025revisiting}, EASE~\cite{zhou2024expandable}, InfLoRA~\cite{liang2024inflora}, SEMA~\cite{wang2025self}, ACMap~\cite{fukuda2025adapter} and CL-LoRA~\cite{he2025cl}.

\noindent \textbf{Training details:} 
We adopt ViT/B-16~\cite{dosovitskiy2020image} pretrained on ImageNet-21K~\cite{deng2009imagenet} as the backbone for all experiments. 
All models are optimized using SGD and 
a cosine learning rate scheduler is applied throughout training. 
The number of training epochs is dynamically adjusted according to the task size. 
For baseline methods, we strictly follow their original training settings for fair comparison.

\noindent \textbf{Evaluation Metrics: }
Following the evaluation protocol of~\cite{rebuffi2017icarl}, we let $A_{t}$ denote the accuracy on all seen classes after learning task $t$. We report the last step accuracy $A_{T}$, which evaluates the performance across the entire class set after the last task $T$, 
as well as the average accuracy $\bar{A}$ across all incremental steps.
In addition, we report the forgetting metric $F$~\cite{chaudhry2018riemannian} which measures the average performance degradation on previous tasks after subsequent learning.

\begin{table*}[t]
\resizebox{\textwidth}{!}{%
\begin{tabular}{l|cc|cc|cc|cc}
\Xhline{0.8pt}
\multirow{2}{*}{\textbf{Method}} 
& \multicolumn{2}{c|}{CIFAR100 (T=10)} 
& \multicolumn{2}{c|}{CUB (T=20)} 
& \multicolumn{2}{c|}{ImageNet-A (T=10)} 
& \multicolumn{2}{c}{ImageNet-R (T=40)} \\
& $A_{T}$ & $\bar{A}$ 
& $A_{T}$ & $\bar{A}$ 
& $A_{T}$ & $\bar{A}$ 
& $A_{T}$ & $\bar{A}$ \\
\hline
L2P~\cite{wang2022learning}           
& 64.63 ± 1.33 & 76.69 ± 1.02 
& 53.04 ± 1.35 & 69.00 ± 1.08 
& 34.52 ± 2.40 & 46.51 ± 3.08 
& 45.79 ± 1.48 & 58.26 ± 1.48 \\

DualPrompt~\cite{wang2022dualprompt}    
& 71.43 ± 1.35 & 81.12 ± 0.46 
& 53.76 ± 0.72 & 70.28 ± 1.41 
& 32.08 ± 2.29 & 42.59 ± 2.64 
& 45.13 ± 1.42 & 55.25 ± 1.19 \\

CODA-Prompt~\cite{smith2023coda}   
& 81.40 ± 0.61 & 87.55 ± 0.34 
& 66.02 ± 1.70 & 77.08 ± 0.76 
& 44.37 ± 0.90 & 53.46 ± 1.41 
& 53.18 ± 1.75 & 62.13 ± 0.79 \\

Aper-adapter~\cite{zhou2025revisiting}  
& 86.47 ± 1.87 & 91.04 ± 2.41 
& 87.31 ± 0.22 & 91.38 ± 0.82 
& 50.74 ± 1.84 & 61.31 ± 3.04 
& 54.61 ± 0.14 & 63.36 ± 1.58 \\

EASE~\cite{zhou2024expandable}          
& \textcolor{blue}{\textbf{86.90 ± 0.73}} 
& \textcolor{blue}{\textbf{91.71 ± 0.88}} 
& 83.12 ± 0.23 & 87.05 ± 2.75 
& 56.11 ± 0.54 & 64.52 ± 1.03 
& 67.02 ± 0.31 & 72.98 ± 0.83 \\

InfLoRA~\cite{liang2024inflora}       
& 83.78 ± 1.01 & 89.24 ± 0.64 
& 56.09 ± 1.05 & 77.22 ± 0.82 
& 46.39 ± 1.75 & 58.40 ± 1.95 
& 60.91 ± 1.63 & 70.03 ± 1.82 \\

SEMA~\cite{wang2025self}          
& 86.53 ± 0.94 & 90.96 ± 0.64 
& 71.83 ± 1.36 & 82.61 ± 1.90 
& 41.16 ± 6.99 & 55.27 ± 4.39 
& 57.56 ± 6.05 & 67.16 ± 5.13 \\

ACMap~\cite{fukuda2025adapter}          
& 84.61 ± 0.19 & 90.27 ± 0.50 
& \textcolor{blue}{\textbf{87.34 ± 0.27}} 
& \textcolor{blue}{\textbf{91.45 ± 0.51}} 
& 50.47 ± 0.12 & 61.61 ± 0.54 
& 60.54 ± 1.45 & 69.13 ± 1.77 \\

CL-LoRA~\cite{he2025cl}       
& 80.06 ± 3.20 & 88.83 ± 1.44 
& 65.48 ± 3.66 & 79.71 ± 2.75 
& \textcolor{blue}{\textbf{56.27 ± 2.29}} 
& \textcolor{blue}{\textbf{66.46 ± 2.45}} 
& \textcolor{blue}{\textbf{69.32 ± 1.12}} 
& \textcolor{blue}{\textbf{76.12 ± 1.61}} \\

\hline
One-A (Ours) 
& \textcolor{red}{\textbf{88.23 ± 0.48}} 
& \textcolor{red}{\textbf{92.22 ± 0.81}} 
& \textcolor{red}{\textbf{88.31 ± 0.40}} 
& \textcolor{red}{\textbf{92.63 ± 0.76}} 
& \textcolor{red}{\textbf{58.28 ± 1.04}} 
& \textcolor{red}{\textbf{69.05 ± 0.57}} 
& \textcolor{red}{\textbf{69.94 ± 0.84}} 
& \textcolor{red}{\textbf{76.97 ± 0.95}} \\
\Xhline{0.8pt}
\end{tabular}
}
\caption{Last task accuracy ($A_{T}$) and average accuracy ($\bar{A}$) on multiple datasets with the step imbalance factor $\gamma=0.01$ and different task numbers. All results are averaged over 5 runs with mean ± standard deviation. \textcolor{red}{\textbf{Red}} denotes the best result and \textcolor{blue}{\textbf{Blue}} denotes the second best in each column.}
\label{tab:AA}
\centering
\end{table*}

\subsection{Benchmark Comparison}
\label{sec:benchmark}
Table~\ref{tab:AA} reports the last-step accuracy ($A_T$) and average accuracy ($\bar{A}$) under a step-imbalance ratio $\gamma = 0.01$.
Our approach achieves the highest accuracy across all datasets while maintaining a single adapter during inference, outperforming multi-adapter methods like EASE and CL-LoRA that require executing multiple task-specific adapters sequentially.
Compared with ACMap, our asymmetric subspace fusion yields more stable and discriminative merging. 
On ImageNet-A and ImageNet-R, our method achieves consistent improvements of 7.8\% and 9.4\% in $A_T$.
These results demonstrate that our method achieves both high efficiency and strong adaptability under step-imbalanced conditions.

\begin{figure*}[t]
    \centering
    \includegraphics[width=\linewidth]{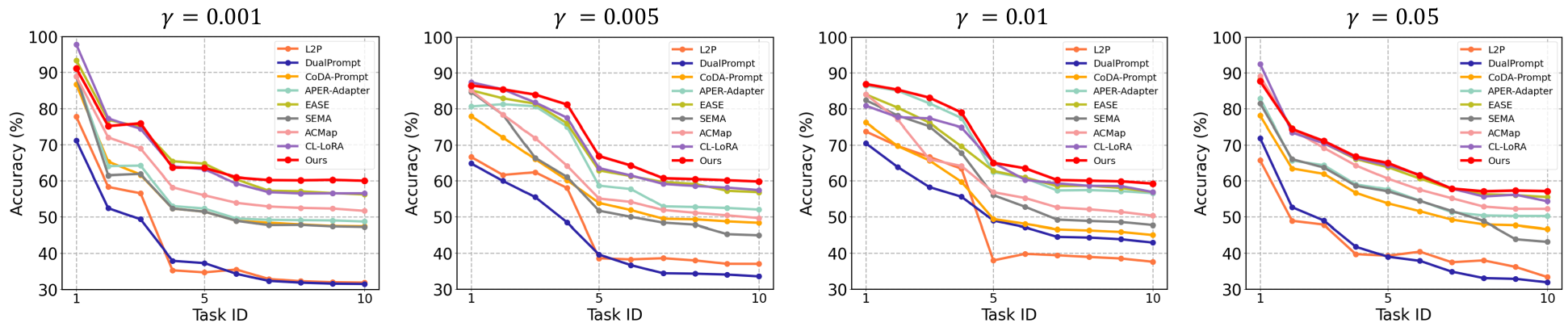}
    \caption{Accuracy at each step on ImageNet-A under different step-imbalance ratios $\gamma \in \{0.001, 0.005, 0.01, 0.05\}$.}
    \label{fig:imbalance_rho}
\end{figure*}

Table~\ref{tab:rho_T} extends the comparison to diverse imbalance regimes and task lengths.
Across both severe ($\gamma=0.001$) and mild ($\gamma=0.05$) settings, and for both $T=10$ and $T=20$, our method achieves the competitive average accuracy and last-task accuracy.
The advantage becomes more pronounced under longer task streams and stronger imbalance, where cumulative interference is more severe and early-task samples are heavily underrepresented.
These results indicate that the proposed subspace fusion generalizes across different imbalance intensities without relying on multiple task-specific adapters. 
Results under a mixed long-tailed and step-imbalance setting are provided in Section~\ref{sec:ltsicil}, where One-A remains competitive while maintaining its inference efficiency.
Figure~\ref{fig:imbalance_rho} further provides a step-wise view of the performance on ImageNet-A under varying imbalance ratios.
As $\gamma$ decreases, all methods show an expected decline in performance due to the stronger dominance of small-task samples and reduced representation of early classes.
Nevertheless, our method achieves comparatively higher accuracy under different imbalance settings, showing clear advantages in mitigating forgetting and preserving performance even under extreme imbalance. 
When $\gamma$ is small, methods like L2P and DualPrompt suffer from rapid degradation because their prompt or key memory modules overfit to the dominant new-task distribution.
Our model sustains a larger gap especially in later tasks, indicating that the direction-specific gating and adaptive preservation mechanism effectively control how much prior subspace is retained as task imbalance intensifies.

\begin{table}[t]
\resizebox{0.45\textwidth}{!}{%
\begin{tabular}{l|cc|cc|cc|cc}
\Xhline{0.8pt}
\multirow{3}{*}{\textbf{Method}} 
& \multicolumn{4}{c|}{$\gamma=0.001$} 
& \multicolumn{4}{c}{$\gamma=0.05$} \\
& \multicolumn{2}{c|}{$T=10$} 
& \multicolumn{2}{c|}{$T=20$} 
& \multicolumn{2}{c|}{$T=10$} 
& \multicolumn{2}{c}{$T=20$} \\
& $A_{T}$ & $\bar{A}$ 
& $A_{T}$ & $\bar{A}$ 
& $A_{T}$ & $\bar{A}$ 
& $A_{T}$ & $\bar{A}$ \\
\hline

Aper-adapter~\cite{zhou2025revisiting}  
& 84.91 & 90.31 
& 82.41 & 87.68 
& 86.32 & 90.34 
& 83.05 & 88.32 \\

EASE~\cite{zhou2024expandable}          
& 87.52 & 91.98
& 84.79 & 90.42 
& \textbf{87.02} & \textbf{91.74}
& 84.42 & 89.23 \\

SEMA~\cite{wang2025self}          
& 84.70 & 88.58 
& 74.85 & 81.86
& 85.63 & 91.18
& 81.48 & 87.85 \\

ACMap~\cite{fukuda2025adapter}          
& 83.43 & 90.15 
& 83.80 & 88.84 
& 85.81 & 91.21 
& 84.25 & 89.08 \\

CL-LoRA~\cite{he2025cl}       
& 82.34 & 89.98
& 80.43 & 89.44
& 83.96 & 89.79
& 76.76 & 84.39 \\

\hline
One-A (Ours) 
& \textbf{88.32} &  \textbf{92.54}
& \textbf{86.27} &  \textbf{90.78}
& 86.83 &  91.46
& \textbf{84.68} &  \textbf{89.58}\\
\Xhline{0.8pt}
\end{tabular}
}
\caption{Last task accuracy ($A_{T}$) and average accuracy ($\bar{A}$) on multiple datasets with different step imbalance factors $\gamma$ and task numbers $T$.}
\label{tab:rho_T}
\centering
\end{table}

\subsection{Ablation Study}
We conduct ablation experiments to investigate how each component contribute to the performance. In particular, we assess (1) the impact of each key component on model stability and adaptability, and (2) the effectiveness of our merging mechanism in balancing performance and efficiency against existing alternatives.

\begin{figure}[t]
\centering
\includegraphics[width=\columnwidth]{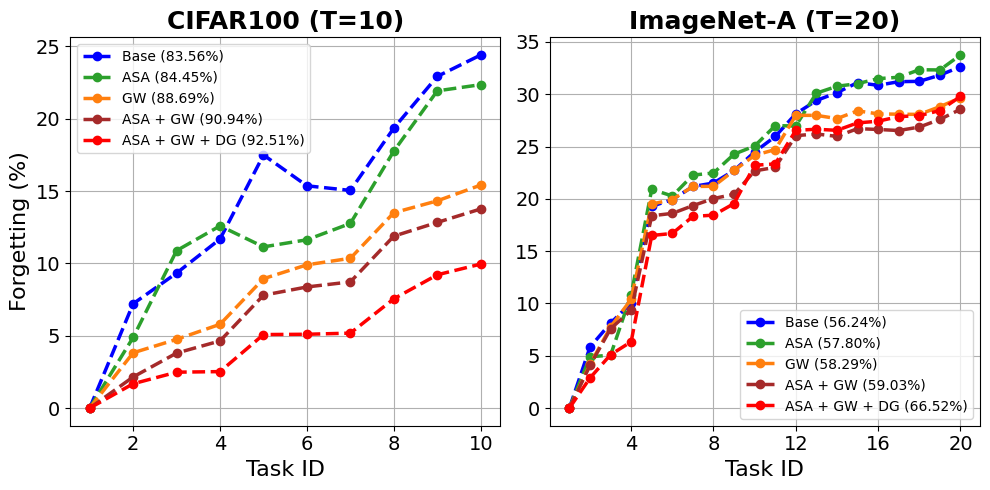}
\captionof{figure}{Forgetting curves under different configurations, along with the corresponding $\bar{A}$.}
\label{fig:forgetting}
\end{figure}

\paragraph{Effectiveness of Individual Components.} We conduct an ablation study to evaluate the contribution of each proposed component, namely Asymmetric SVD Alignment (ASA) introduced in Section~\ref{subsec:SAS}, Information-Adaptive Global Weighting (GW) described in Section~\ref{subsec:GW}, and Directional Gating (DG) presented in Section~\ref{subsec:DG}. 
Under a step-imbalance ratio of $\gamma = 0.01$, the results on CIFAR-100 and ImageNet-A with different task numbers are summarized in Table~\ref{tab:ablation}.
When ASA is removed, the model degenerates into the symmetric KnOTS-style SVD merging, where all task updates are treated equally, leading to noticeable performance degradation due to subspace drift from small-task noise. 
Replacing GW with naive averaging also reduces performance, confirming the necessity of adaptively weighting tasks according to their information content. 
Finally, the removal of DG, which controls the fusion strength per singular direction, results in less stable updates and weaker adaptation, especially under larger task streams. 
Overall, each component contributes positively to the final performance, and their combination yields the most balanced trade-off between stability and plasticity across both benchmarks.

To provide a more fine-grained view of knowledge retention, Figure~\ref{fig:forgetting} shows the task-wise forgetting curves on CIFAR-100 (T=10) and ImageNet-A (T=20) along with the corresponding average accuracies.
On CIFAR-100, the effect is particularly pronounced: ASA and GW each lower the forgetting trajectory across most tasks, and their combination further suppresses the accumulation of forgetting.
With DG enabled, the full model achieves the lowest forgetting throughout the entire sequence, while also yielding the highest average accuracy, indicating a better stability–plasticity trade-off.
On the longer ImageNet-A stream, GW reduces forgetting compared to the Base model, while their combination further improves retention across most tasks.
When DG is added, the forgetting curve becomes slightly higher than that of ASA+GW in the later stages, but this is accompanied by a substantial gain in average accuracy.
This indicates a clear stability–plasticity trade-off, where DG allows more adaptation to new tasks at the cost of mildly increased forgetting, ultimately leading to better overall performance.


\begin{table}[t]
\centering
\resizebox{0.9\columnwidth}{!}{%
\begin{tabular}{ccc|cccc}
\hline
\multirow{2}{*}{\textbf{ASA}} & \multirow{2}{*}{\textbf{GW}} & \multirow{2}{*}{\textbf{DG}} & \multicolumn{2}{c}{\textbf{CIFAR-100}} & \multicolumn{2}{c}{\textbf{ImageNet-A}} \\
 & & & $T=10$ & $T=20$ & $T=10$ & $T=20$ \\
\hline
 & &                                    & 72.43 & 78.34& 45.82&47.60 \\
$\checkmark$ & &                        & 75.82 & 81.27& 48.02&48.39 \\
 & $\checkmark$ &                       & 82.46& 80.74& 51.32&49.24 \\
$\checkmark$ & $\checkmark$ &           & 84.31& 82.76& 53.65&52.01 \\
$\checkmark$ & $\checkmark$ & $\checkmark$ &\textbf{ 88.39} & \textbf{85.32}& \textbf{59.25}&\textbf{56.62} \\
\hline
\end{tabular}%
}
\caption{Ablation study of each component of One-A on CIFAR100 and ImageNet-A.}
\label{tab:ablation}
\end{table}

\begin{table}[t]
\centering
\resizebox{0.9\columnwidth}{!}{%
\begin{tabular}{lcccc}
\toprule
Method & $A_T$ (\%) & $w\bar{A}$ (\%) & FLOPs & \#Adapters \\
\midrule
Per-task adapter &  43.15  &  55.67  &  $39.19\times$ & $40 (T)$ \\
MOS~\cite{sun2025mos}        &  67.73  &  74.04  &  $40.60\times$ & $40 (T)$ \\
ACMap~\cite{fukuda2025adapter}          &  61.32  &  68.56  &  $0.99\times$ & 1 \\
\midrule
One-A (Ours)                 &  \textbf{70.55}  &  \textbf{76.95}  &  $1.\times$ & 1 \\
\bottomrule
\end{tabular}
}
\caption{Comparison of merging strategies on SI-CIL. }
\label{tab:ablation_fusion}
\end{table}

\paragraph{Comparison of Merging Strategies.} To evaluate the effectiveness and efficiency of our merging strategy, we conduct experiments on the ImageNet-R dataset with $T{=}40$ incremental tasks and a step-imbalance ratio of $\gamma{=}0.01$. We compare our method with three alternatives: the per-task adapter, which keeps an independent adapter for each task without merging, MOS~\cite{sun2025mos}, which merges adapters by optimizing a scalar interpolation coefficient, and ACMap~\cite{fukuda2025adapter}, which performs iterative averaging–based adapter merging.
We report the last-step accuracy ($A_T$), weighted average accuracy ($\bar{wA}$), the FLOPs ratio at inference, and the number of adapters required.

As shown in Table~\ref{tab:ablation_fusion}, merging adapters consistently improves accuracy over keeping per-task adapters, indicating that consolidating cross-task information mitigates interference and enhances generalization.
Among the merging baselines, MOS achieves moderate improvement by effectively alleviating inter-task interference through its EMA-based adapter merging. However, during inference, each sample must traverse all task-specific adapters and perform an additional similarity-based retrieval, causing the inference cost to grow linearly with the number of tasks. This design leads to substantially higher FLOPs (${\sim}40\times$ ours) and makes the approach less practical for long task sequences.
ACMap in contrast maintains a single adapter through weight averaging and achieves very low inference cost. However, it lacks mechanisms to handle subspace misalignment or task-size disparity, leading to degraded robustness under our step-imbalanced scenario.
In comparison, our method not only achieves the highest accuracy but also preserves the efficiency of single-adapter inference. By performing asymmetric subspace alignment during fusion, our approach effectively preserves dominant directions from large tasks while integrating small-task knowledge in a stable manner. This yields superior balance between adaptivity and efficiency, achieving both faster inference and higher robustness than ACMap, and substantially lower computational cost than MOS.

\subsection{Discussion}

\begin{figure}[t]
    \centering
    \begin{subfigure}{0.48\linewidth}
        \includegraphics[width=\linewidth]{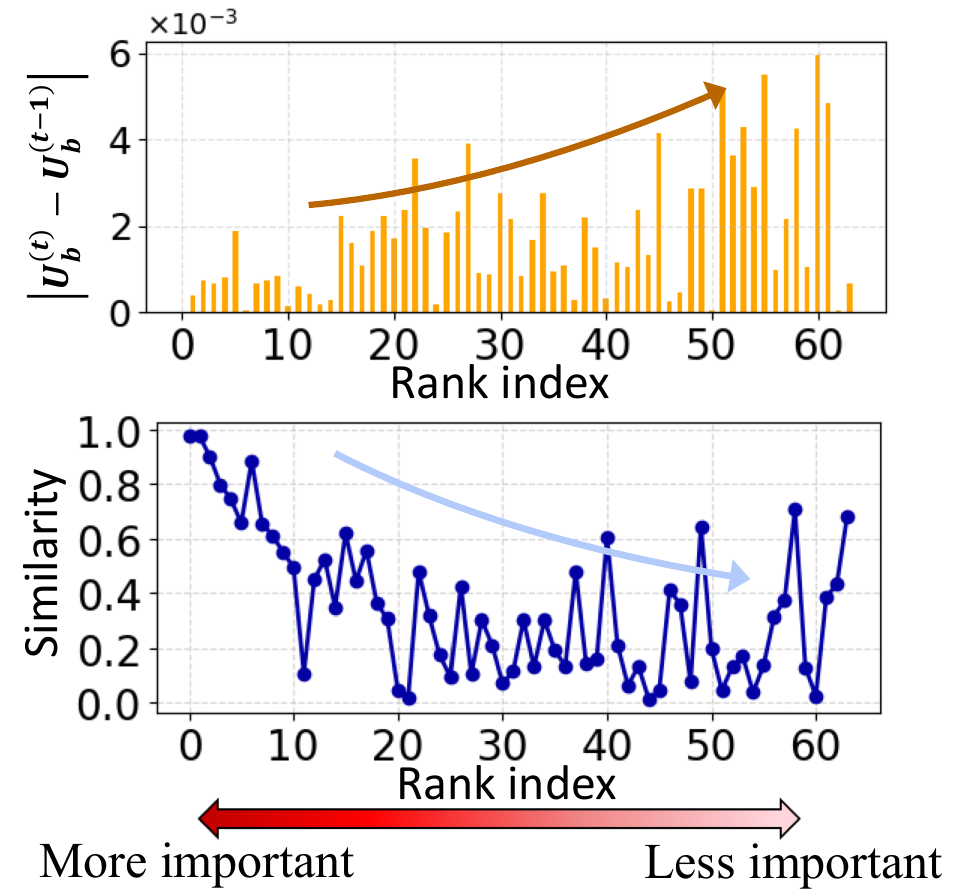}
        \caption{Quantitative analysis of directional updates.}
        \label{fig:rank_a}
    \end{subfigure}
    \hspace{0.02\linewidth}
    \begin{subfigure}{0.48\linewidth}
        \includegraphics[width=\linewidth]{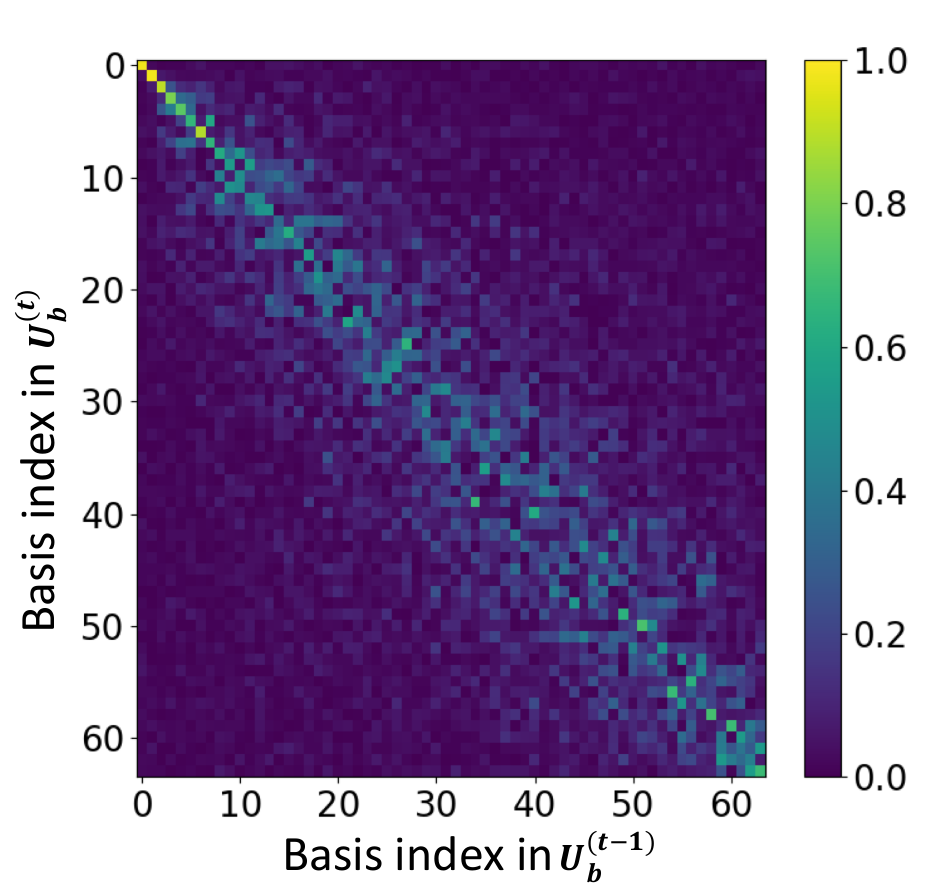}
        \caption{Visual comparison of basis alignment.}
        \label{fig:rank_b}
    \end{subfigure}
    \caption{Illustration of the subspace behavior during adapter merging.}
    \label{fig:rank}
\end{figure}

\begin{figure}
\centering
\begin{subfigure}{0.4\linewidth}
    \includegraphics[width=\linewidth]{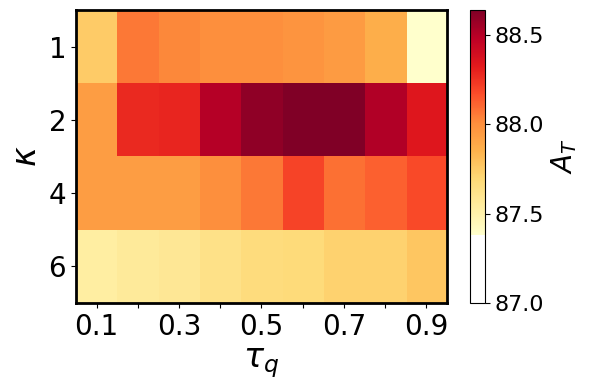}
    \caption{CIFAR-100}
    \label{fig:gating_cifar}
\end{subfigure}
\hspace{0.02\linewidth}
\begin{subfigure}{0.4\linewidth}
    \includegraphics[width=\linewidth]{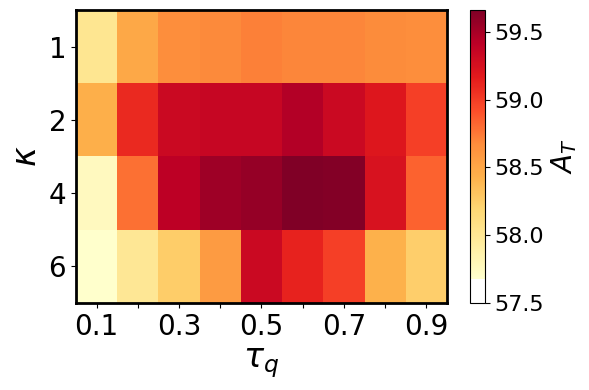}
    \caption{ImageNet-A}
    \label{fig:gating_ina}
\end{subfigure}
\caption{Sensitivity analysis on the gating parameters. Last-step accuracy ($A_T$) under different quantile threshold $\tau_q$ and gating sharpness $\kappa$.}
\label{fig:gating_sensitivity_reduced}
\end{figure}

\paragraph{Directional update and basis alignment analysis.} 
Figure~\ref{fig:rank} illustrates how asymmetric adapter merging affects the internal subspace structure through both quantitative and visual analyses.
Figure~\ref{fig:rank_a} presents the quantitative analysis of directional updates across singular directions. 
Overall, directions with larger rank indices, which correspond to less important basis in the base adapter, tend to receive stronger updates. 
This behavior enables the model to incorporate new information while preserving the stability of the principal subspace that represents old knowledge. 
However, the trend is not strictly monotonic. 
The actual update magnitude depends on both the gating weight and the activation strength of each direction under the new task. 
Consequently, some head directions with small gating coefficients may still undergo noticeable adjustments when they are highly activated, whereas uninformative tail directions remain nearly unchanged even with larger weights. 
This adaptive interaction forms a natural balance between stability and plasticity, allowing the gating mechanism to focus updates on task-relevant components without imposing a rigid head–tail separation.
Figure~\ref{fig:rank_b} provides a visual comparison of basis alignment before and after merging. 
The heatmap shows strong diagonal dominance with mild dispersion in the tail region, indicating that the principal subspace of the base adapter is well preserved while minor bases are softly rotated to integrate new information. 
Together, these findings suggest that our asymmetric fusion achieves direction-aware preservation and maintains the stability of major representations while reallocates capacity for new task learning under SI-CIL.

\paragraph{Sensitivity Analysis of the Direction Gating.}
Figure~\ref{fig:gating_sensitivity_reduced} visualizes the sensitivity of our direction-specific gating mechanism (Section~\ref{subsec:DG}) to the quantile threshold $\tau_q$ and the gating sharpness $\kappa$ on CIFAR100 and ImageNet-A when $T=10$ and $\gamma=0.01$. 
The quantile threshold $\tau_q$ determines the split between the head and tail singular directions, where the head region primarily preserves previous knowledge and the tail region accommodates new information. 
The parameter $\kappa$ controls how sharply the gating function transitions between these two regions, with larger values yielding steeper separation. 
The heatmap shows that overall performance remains stable within a narrow range, but moderate values of $\tau_q$ and $\kappa$ consistently yield the best results. 
Extremely small or large thresholds lead to either excessive rigidity or overfitting to new tasks, while overly sharp transitions reduce the smoothness of knowledge transfer across directions. 
These results suggest that the gating mechanism effectively regulates the trade-off between stability and plasticity by controlling how the update strength is distributed across singular directions.

\vspace{-3mm}
\section{Conclusion}
\vspace{-2mm}
In this work, we revisit class-incremental learning (CIL) under the step–imbalanced scenario, where tasks arrive with highly uneven class scales. This setting better reflects real-world incremental learning compared to balanced CIL, but exposes severe challenges such as optimization bias, subspace distortion, and unstable update. To address these issues, we propose an asymmetric subspace alignment framework that extends model merging to MLP adapters and explicitly accounts for task-size disparity during fusion. Our method preserves dominant directions from large tasks while efficiently injecting new knowledge from smaller ones, ensuring both stability and adaptability across diverse task streams. Extensive experiments on multiple CIL benchmarks demonstrate that our approach achieves consistently superior accuracy and efficiency under step-imbalanced conditions, validating its effectiveness and scalability for realistic continual learning.

{
    \small
    \bibliographystyle{ieeenat_fullname}
    \bibliography{main}
}

\clearpage
\setcounter{page}{1}
\maketitlesupplementary

\section*{Overview} 
This supplementary material provides additional details and results for One-A:
\begin{itemize}
    \item \textbf{Section~\ref{sec:more_sicil}} gives a detailed formulation of step-imbalanced class-incremental learning (SI-CIL), including how exponential class ratios are generated and converted into head–tail task sequences, and further clarifies the differences between SI-CIL and long-tailed CIL.
    \item \textbf{Section~\ref{sec:extend}} elaborates on the One-A fusion procedure, including pseudo code in Section~\ref{subsec:code} and clarification on the exemplar-free setting, the pre-trained paradigm in Section~\ref{subsec:ptm_paradigm}, the limitations of symmetric alignment in Section~\ref{subsec:knot_limitation}, variants of the information proxy in Section~\ref{subsec:proxy_variants}, the adaptive training strategy for step-imbalanced tasks in Section~\ref{subsec:adaptive_training}, and a complexity analysis in Section~\ref{subsec:com_com}.
    \item \textbf{Section~\ref{sec:more_exp}} reports additional experimental details and ablations, including detailed implementation settings in Section~\ref{subsec:detail_setting}, results under the descending step-imbalanced scenarios in Section~\ref{subsec:descending}, performance in the balanced setting in Section~\ref{subsec:balance}, adapter rank sensitivity in Section~\ref{subsec:adapter_rank}, and inference efficiency in Section~\ref{subsec:inference_eff}.
    \item \textbf{Section~\ref{sec:ltsicil}} reports additional results under a mixed long-tailed and step-imbalance setting and discusses future directions.
\end{itemize}

\noindent \textbf{The source code will be made publicly available.}

\section{More Description about Step-Imbalanced Class-Incremental Learning}
\label{sec:more_sicil}

Unlike long-tailed classification, where classes have unequal numbers of samples, SI-CIL keeps the number of samples per class fixed but varies the number of classes across tasks. This results in tasks of highly uneven scale, ranging from large head tasks to extremely small tail tasks.

\begin{figure}[t]
    \centering
    \includegraphics[width=0.9\columnwidth]{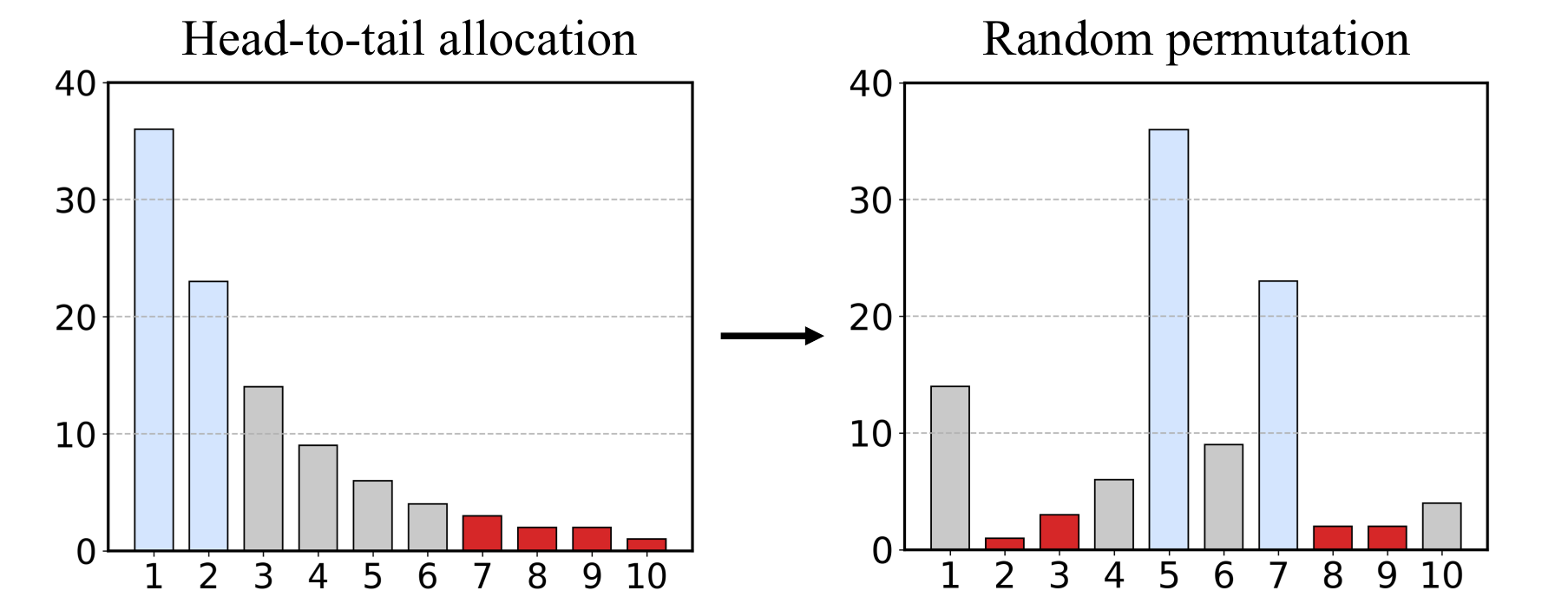}
    \caption{Generation of the step-imbalanced task sequence. 
    \textbf{Left: }head-to-tail allocation based on exponential class ratios. 
    \textbf{Right: }random permutation producing the final task order.}
    \label{fig:step_imbalance_generation}
\end{figure}

To construct the step-imbalanced distribution, we begin with $C$ classes split into $T$ tasks and generate an exponential class ratio 
\begin{equation}
r_k = \gamma^{\frac{k}{C-1}}, \qquad k = 0,\ldots,C-1,
\label{eq:im_ratio}
\end{equation}
where $\gamma$ is the step-imbalance factor. After normalization, these ratios determine how many classes are assigned to each task in a head-to-tail manner, producing large tasks followed by progressively smaller tasks, as illustrated on the left side of Figure~\ref{fig:step_imbalance_generation}.
To emulate realistic non-sequential task arrivals, we then randomly permute the task order, yielding the final step-imbalanced sequence shown on the right side of Figure~\ref{fig:step_imbalance_generation}. This process creates a clear distinction between head tasks with rich class diversity and tail tasks with very limited class information.

We consider the above construction in a class-incremental learning (CIL) scenario in which the number of
classes introduced at each incremental step varies substantially across tasks.
Formally, a sequence of $T$ tasks is given as 
$\{\mathcal{D}_1,\ldots,\mathcal{D}_T\}$, where
\[
\mathcal{D}_t=\{(x_i,y_i)\}_{i=1}^{N_t},
\quad
y_i \in \mathcal{Y}_t,
\quad
\mathcal{Y}_t \cap \mathcal{Y}_{t'} = \emptyset\;\text{for }t\neq t'.
\]

\begin{figure*}[t]
    \centering
    \includegraphics[width=0.8\textwidth]{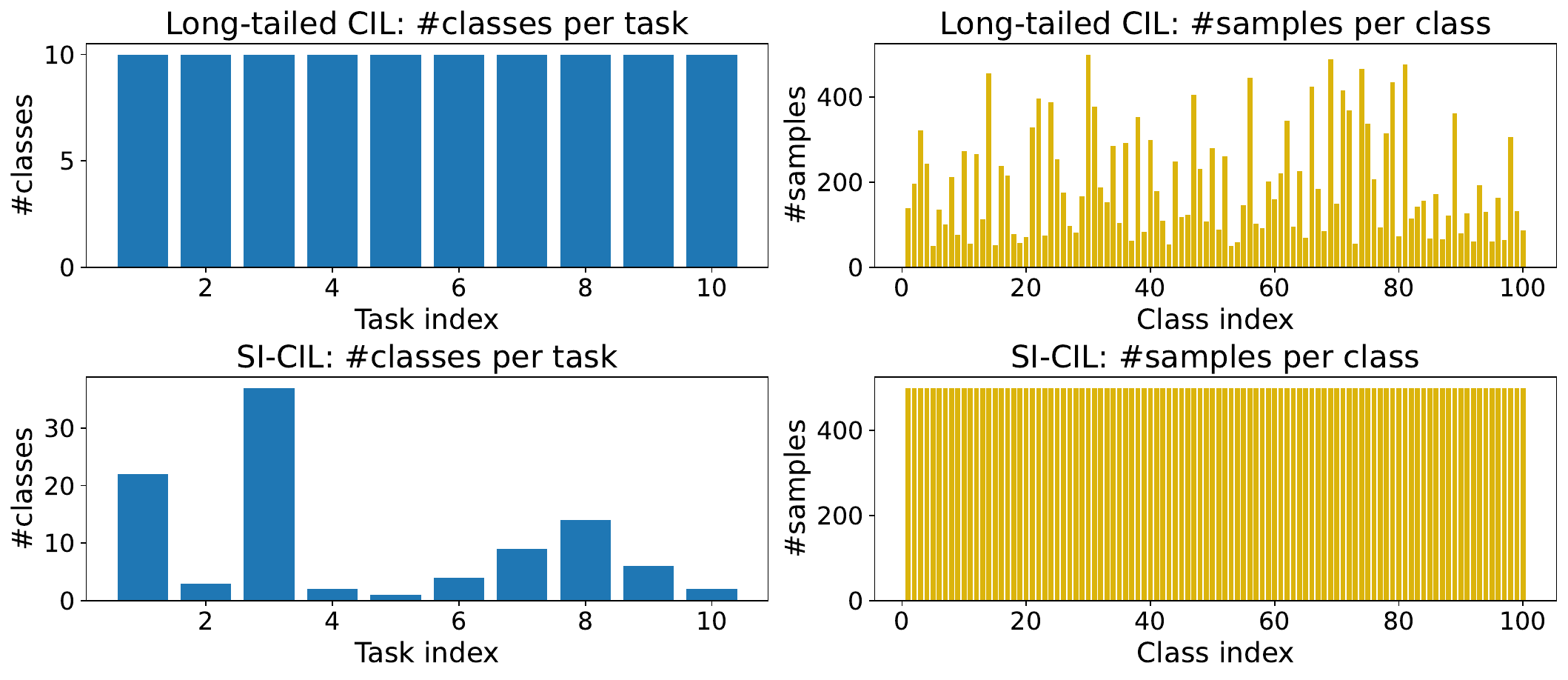}
    \caption{\textbf{Illustration of long-tailed CIL vs. SI-CIL.}
    We show an example with $T{=}10$ tasks and $C{=}100$ classes.
    \textbf{Top (Long-tailed CIL):} each task contains the same number of classes, while the number of samples per class is long-tailed.
    \textbf{Bottom (SI-CIL):} the number of samples per class is fixed, while the number of classes varies significantly across incremental steps.}
    \label{fig:sicil_vs_longtail}
\end{figure*}

\paragraph{\textbf{SI-CIL vs. Long-tailed CIL.}}
Different from conventional CIL where each incremental step introduces a fixed number of classes, 
step-imbalanced class-incremental learning (SI-CIL) allows the class count $|\mathcal{Y}_t|$ to vary substantially across tasks, resulting in a strongly step-imbalanced task stream.
Throughout this setting, the number of samples per class is kept constant, so the imbalance arises purely from differences in task-level class counts rather than sample frequency.
Figure~\ref{fig:sicil_vs_longtail} illustrates the conceptual difference between SI-CIL and long-tailed CIL using an example with $T{=}10$ tasks and $C{=}100$ classes.
In long-tailed CIL, each task typically contains a fixed number of classes (left-top), while the number of samples per class is highly imbalanced and follows a long-tailed distribution (right-top).
In contrast, SI-CIL fixes the number of samples per class (right-bottom) but introduces imbalance at the task level: different incremental steps contain vastly different numbers of new classes (left-bottom), leading to head tasks with rich class diversity and tail tasks with extremely limited class information.
This controlled formulation disentangles task-level class-count disparity from sample-level imbalance, and highlights the unique challenges induced by step-size variation.

\section{Extended Description of One-A}
\label{sec:extend}
\subsection{Pseudo Code of One-A}
\label{subsec:code}

Algorithm~\ref{alg:onea_core} details the asymmetric fusion routine used in One-A to merge the new task adapter with the previously accumulated one. At each task, the algorithm selects a base adapter $b$ and an align adapter $a$ according to task size, ensuring that the dominant representational subspace is always defined by the more informative task. For every layer $\ell$, the base adapter $\Delta W_b^{(\ell)}$ is decomposed via SVD,
capturing its principal subspace through $(U_b, \Sigma_b, V_b)$ (Eq.~\ref{eq:base}).
The align adapter is then projected into this subspace using the alignment mapping in Eq.~\ref{eq:adapter_align}, which prevents small-task updates from significantly altering the dominant directions of the base.
The globally fused representation is obtained by combining the two aligned components using the information-adaptive weights of Eq.~\ref{eq:global_weight}, producing the intermediate fusion $V_{\mathrm{fused}}$ as in Eq.~\ref{eq:global_fusion}. To further balance stability and plasticity across singular directions, a direction-wise gating vector $g$ is computed from the normalized singular spectrum of $\Sigma_b$, following Eq.~\ref{eq:directional_weight}. This gating modulates the fusion strength along each direction, reinforcing high-energy directions while allowing the model to incorporate complementary information from low-energy ones as in Eq.~\ref{eq:dirc_fuse}.
Finally, the gated update is reassembled into the merged adapter via Eq.~\ref{eq:directional_fusion}, yielding a single updated adapter after each task. By combining asymmetric subspace alignment, information-adaptive weighting, and energy-aware directional gating, One-A maintains a stable shared representation while flexibly integrating new task information 
under severe step-size imbalance without increasing inference-time complexity.

\paragraph{Clarification on the Exemplar-Free Setting.}
As illustrated in the overall training procedure in Algorithm~\ref{alg:onea}, One-A strictly follows the exemplar-free protocol: at each task $t$, the model is trained only on the current task data $\mathcal{D}_t$, produces a temporary task adapter $\Delta W_t^{\mathrm{new}}$, merges it with the accumulated adapter, and computes a set of task-level prototypes $\mathcal{P}_t$ based solely on the updated model parameters. No original samples from $\mathcal{D}_1,\ldots,\mathcal{D}_{t-1}$ are ever revisited or reused. It is important to emphasize that the use of these prototypes does not violate the exemplar-free setting. In the continual learning literature, \emph{exemplar} refers to retained \emph{original training samples} like images or raw feature vectors from previous tasks for replay-based learning~\cite{rebuffi2017icarl,yan2021dynamically,douillard2020podnet}. 
In contrast, methods including prompt and adapter based approaches such as DualPrompt~\cite{wang2022dualprompt} and EASE~\cite{zhou2024expandable} are considered as exemplar-free as long as it does not preserve or reuse original past samples during training.
Therefore, One-A fully adheres to the exemplar-free principle.

\begin{algorithm}[t]
\caption{Asymmetric Subspace Fusion in One-A}
\label{alg:onea_core}
\begin{algorithmic}[1]

\REQUIRE Backbone adapter $\{\Delta W_m^{(\ell)}\}_{\ell=1}^L$, new task adapter $\{\Delta W_t^{(\ell)}\}_{\ell=1}^L$, data sizes $\{|\mathcal{D}_i|\}$, information measures $\{\mathrm{Info}_i\}$, information proxy $\phi(\cdot)$
\ENSURE Updated merged adapter $\{\Delta W_m^{(\ell)}\}_{\ell=1}^L$

\IF{$t = 1$}
    \FOR{$\ell = 1$ to $L$}
        \STATE $\Delta W_m^{(\ell)} \leftarrow \Delta W_t^{(\ell)}$   \scright{$\triangleright$ initial task, no fusion}
    \ENDFOR
    \STATE \textbf{return} $\{\Delta W_m^{(\ell)}\}_{\ell=1}^L$
\ENDIF

\IF{$|\mathcal{D}_t| \ge \sum_{i=1}^{t-1} |\mathcal{D}_i|$}
    \STATE $b \leftarrow t$, $a \leftarrow m$ \scright{$\triangleright$ new task is larger}
\ELSE
    \STATE $b \leftarrow m$, $a \leftarrow t$ \scright{$\triangleright$ merged adapter is larger}
\ENDIF

\STATE $w_a \leftarrow \dfrac{\phi(\mathrm{Info}_a)}{\phi(\mathrm{Info}_a)+\phi(\mathrm{Info}_b)}$, 
       $w_b \leftarrow 1 - w_a$ \scright{$\triangleright$ Eq.~\ref{eq:global_weight}}

\FOR{$\ell = 1$ to $L$}

    \STATE \textbf{// Asymmetric SVD on base adapter}
    \STATE $\Delta W_b^{(\ell)} = U_b \Sigma_b V_b^\top$ \scright{$\triangleright$ Eq.~\ref{eq:base}}

    \STATE \textbf{// Project align adapter}
    \STATE $V_{a \rightarrow b} \leftarrow \Delta W_a^{(\ell)\top} U_b \Sigma_b^{-1}$ \scright{$\triangleright$ Eq.~\ref{eq:adapter_align}}

    \STATE \textbf{// Global fusion}
    \STATE $V_{\mathrm{fused}} \leftarrow w_b V_b + w_a V_{a \rightarrow b}$ \scright{$\triangleright$ Eq.~\ref{eq:global_fusion}}

    \STATE \textbf{// Energy-aware gating}
    \STATE Extract singular values $s_1,\dots,s_r$ from $\Sigma_b$
    \STATE $\tilde{s}_i \leftarrow \dfrac{s_i}{s_1 + \delta}$  for $i=1,\dots,r$  
    \STATE $\theta \leftarrow Q_q(\{\tilde{s}_i\})$
    \STATE $g_i \leftarrow \sigma\!\left(\kappa(\theta - \tilde{s}_i)\right)$
    \STATE $g \leftarrow [g_1,\dots,g_r]^\top$  \scright{$\triangleright$ Eq.~\ref{eq:directional_weight}}

    \STATE \textbf{// Directional fusion}
    \STATE $V_{\mathrm{final}} \leftarrow V_b + g \odot (V_{\mathrm{fused}} - V_b)$  \scright{$\triangleright$ Eq.~\ref{eq:dirc_fuse}}

    \STATE \textbf{// Reconstruct merged adapter}
    \STATE $\Delta W_m^{(\ell)} \leftarrow U_b \Sigma_b V_{\mathrm{final}}^\top$   \scright{$\triangleright$ Eq.~\ref{eq:directional_fusion}}

\ENDFOR

\STATE \textbf{return} $\{\Delta W_m^{(\ell)}\}_{\ell=1}^L$

\end{algorithmic}
\end{algorithm}

\subsection{Pre-trained Paradigm of One-A}
\label{subsec:ptm_paradigm}
In this work, we study Step-Imbalanced Class-Incremental Learning under the paradigm of pre-trained models (PTMs). We follow the standard PTM-based exemplar-free CIL setting to ensure fair comparison with recent state-of-the-art methods \cite{Zhou_2024}.
Most modern CIL approaches build upon strong pre-trained backbones and adapt them to new tasks via lightweight modules such as prompts \cite{wang2022learning,wang2022dualprompt,smith2023coda} or adapters \cite{he2025cl,liang2024inflora,zhou2025revisiting,zhou2024expandable,wang2025self}, while freezing the majority of parameters.
This paradigm has been widely adopted due to its superior stability and efficiency compared to training from scratch.
One-A is an adapter-based method that operates on top of a fixed pre-trained backbone.
Adapters provide a structured and controllable mechanism for continual adaptation in PTMs, making them a natural substrate for studying how task updates should be consolidated under step-size imbalance.
Without a pre-trained backbone, the notion of merging task-specific adaptations into a shared representation becomes ill-defined, as the dominant representational subspace itself would be highly unstable.
We emphasize that the performance gains of One-A should not be attributed to the use of pre-trained models per se, as all compared methods operate under the same PTM-based setting.
Instead, our contribution lies in introducing a previously underexplored learning paradigm, Step-Imbalanced Class-Incremental Learning, 
and systematically analyzing the gap between existing CIL methods and this practical setting. 
Building on this analysis, we propose an efficient and principled adapter-merging mechanism that preserves dominant representational subspaces while integrating minority task updates under severe step-size imbalance.

\subsection{Symmetric Alignment Limitations}
\label{subsec:knot_limitation}
KnOTS~\cite{stoica2024model} is a recent gradient-free and data-free model fusion method that improves over simple parameter averaging by explicitly aligning the task-update subspaces before merging, aiming to address the subspace misalignment commonly observed in LoRA models. Specifically, given $n$ task updates $\{\Delta W^{(1)}, \Delta W^{(2)}, \dots, \Delta W^{(n)}\}$ derived from the same pre-trained model, KnOTS concatenates them into a single matrix $X = \big[\Delta W^{(1)} | \Delta W^{(2)} | \cdots | \Delta W^{(n)}\big]$, and applies singular value decomposition $X = U \Sigma V^\top$, where $U \Sigma$ defines a shared subspace across all tasks and each $V^{(i)}$ captures the task-specific transformation within that shared space. After alignment, KnOTS merges the $V^{(i)}$ components (e.g., by weighted averaging) to obtain a single fused representation $V^{(\mathrm{merged})}$, and reconstructs the final merged update as $\Delta W^{(\mathrm{merged})} = U \Sigma V^{(\mathrm{merged})\top}$. This procedure effectively aligns the task-update subspaces and simplifies downstream merging. Although KnOTS was originally proposed for LoRA updates, the alignment procedure itself only depends on the linear structure of parameter updates rather than LoRA-specific properties. Since MLP adapters also produce structured low-dimensional updates, KnOTS can be directly applied in our setting.

However, KnOTS treats all task updates symmetrically during concatenation, implicitly assuming that they have comparable scale and information content. In step-imbalanced scenarios, small tasks with low information volume still participate equally in defining the shared subspace $U\Sigma$. As a result, both large and small tasks are effectively pulled toward an intermediate subspace that lies between their respective update directions. This compromise forces all tasks to deviate from their original subspaces, but the cost is uneven: large tasks, which contribute richer and more stable dominant directions, lose more critical information than they gain from alignment with small tasks. This asymmetry leads to subspace rotation or drift dominated by small-task noise, ultimately degrading the representational quality of the fused model.

This effect is also quantitatively verified by the ablation results in Table~\ref{tab:ablation}.
A comparison between Rows 1 and 2 (no ASA $\rightarrow$ with ASA) as well as between Rows 3 and 4 (GW only $\rightarrow$ GW + ASA) consistently demonstrates that incorporating asymmetric subspace alignment yields notable accuracy gains. 
These patterns highlight that symmetric alignment treats all tasks equally, which may not be ideal in step-imbalanced settings. Our asymmetric mechanism protects the majority task information by fixing its dominant subspace and injects minority updates only in a controlled way that avoids harming the main representation. ASA stabilizes the dominant subspace by anchoring it to the high-information task (Eq.~\ref{eq:base}), ensuring that small-task projections remain constrained (Eq.~\ref{eq:adapter_align}) and cannot significantly distort the principal components. This empirically validates our hypothesis that symmetric fusion is insufficient under step-size imbalance and highlights ASA as a crucial component of One-A.

\subsection{Variants of Information Proxy}
\label{subsec:proxy_variants}
In the main paper, we adopt the number of classes in each task as the information proxy 
$\phi(\mathrm{Info}_t)$ for computing the global fusion weights (Eq.~\ref{eq:weight_class}). This proxy is simple and stable, and it provides an effective estimate of task diversity in SI-CIL, where the dominant variability mainly comes from the disparity in class counts. 
Moreover, this proxy avoids additional noise from small-task updates and does not require extra computation, which makes it particularly suitable for post-hoc adapter fusion. Nevertheless, the One-A framework is fully compatible with alternative information measures.  
Below we experimented with two representative variants based on the magnitude and structure of the adapter updates.

\paragraph{Norm-based information proxy.}
Frobenius energy (i.e., $\|\cdot\|_F^2$) is a standard measure of matrix information. 
Given an update matrix $\Delta W^{(\ell)}$ at layer $\ell$, we define the norm-based proxy as
\begin{equation}
\phi(\mathrm{Info})
= \big\|\Delta W^{(\ell)}\big\|_F^{\gamma}
= \left( \sum_{i,j} \big(\Delta W^{(\ell)}_{ij}\big)^2 \right)^{\gamma/2},
\end{equation}
where $\gamma > 0$ controls the emphasis on high-energy updates.
In the following experiment, we set $\gamma=1$, which provides a linear and unbiased estimate of update magnitude without amplifying or compressing the contribution from different tasks.
This proxy captures the overall magnitude of the parameter changes and provides a coarse estimate of task informativeness.

\paragraph{Singular-value-based information proxy.}
Another proxy is based on the singular value spectrum of the update.  
Given the SVD
\begin{equation}
\Delta W^{(\ell)} = U \Sigma V^\top,
\end{equation}
with singular values $\{s_i\}$ on the diagonal of $\Sigma$, we define
\begin{equation}
\phi(\mathrm{Info})
= \|\Sigma\|_F^2
= \sum_i s_i^2,
\end{equation}
which measures the total singular value energy of the update.

\begin{table}[t]
\centering
\resizebox{0.8\columnwidth}{!}{%
\begin{tabular}{lcc}
\toprule
Method & CIFAR100 & ImageNet-A \\
\midrule
Norm-based & 86.23 & 55.17 \\
SV-based & 85.83 & 53.65 \\
Class count-based & 88.58 & 59.64 \\
\bottomrule
\end{tabular}
}
\caption{Comparison of information proxies for global fusion in One-A under the SI-CIL setting. Results are last-step accuracy (\%) on CIFAR100 and ImageNet-A.}
\label{tab:info}
\end{table}

Table~\ref{tab:info} shows that all three information proxies are feasible within the One-A framework, but their behaviors differ noticeably in step-imbalanced settings.  
The norm-based and SV-based proxies require additional computation and tend to be more sensitive to noisy or low-rank updates from small tasks, which can lead to less stable fusion, particularly on ImageNet-A.  
In contrast, the class-count proxy yields the highest accuracy on both benchmarks, suggesting that this simple task-aware measure provides a more reliable estimate of task diversity under SI-CIL.  
For these reasons, we adopt the class-count proxy in the main paper.

\subsection{Adaptive Training Strategy for Step-Imbalanced Tasks}
\label{subsec:adaptive_training}
To improve representation learning under step-imbalanced CIL, we introduce a contrastive loss as an auxiliary objective. While the cross-entropy loss $\mathcal{L}_{\mathrm{CE}}$ aligns logits with class labels, it does not impose explicit geometric constraints on the feature space. This limitation is particularly severe for small tasks with only a few classes, where the learned representations tend to be unstable or easily collapsed. In the extreme case of a single-class task, the cross-entropy loss provides no meaningful supervision, and the adapter is therefore left unchanged. To address this issue, we adopt the contrastive loss~\cite{1640964} that explicitly encourages intra-class compactness and inter-class separation. Given $\ell_2$-normalized features $\{f_i\}$ and labels $y_i$, the contrastive loss is defined as
\begin{equation}
\mathcal{L}_{\mathrm{ctr}}
=
\left(1 - \mathrm{sim}(f_i, f_j)\right)
+
\max\!\bigl(0,\ \mathrm{sim}(f_i, f_k) - \tau \bigr),
\label{eq:contra_loss}
\end{equation}
where $(i,j)$ denotes a positive pair with $y_i = y_j$, $(i,k)$ denotes a negative pair with $y_i \ne y_k$, $\mathrm{sim}(\cdot,\cdot)$ is cosine similarity, and $\tau$ is a margin.

Because small tasks benefit disproportionately from such structural guidance, we
introduce a task-size–adaptive weighting scheme that scales the contrastive
regularization according to the number of classes in each incremental step.  
Let $|\mathcal{Y}_t|$ denote the number of classes introduced in task $t$.
The weighting in Eq.~\ref{eq:loss} is applied as
\begin{equation}
\lambda(t)
=
\lambda_{\min}
+
\bigl(\lambda_{\max}-\lambda_{\min}\bigr)
\exp\!\bigl(-k \, (|\mathcal{Y}_t|-1)\bigr),
\label{eq:contra_loss_final}
\end{equation}
so that tasks with fewer classes receive a larger contrastive contribution,
whereas tasks rich in class diversity rely primarily on cross-entropy
supervision.

To further balance optimization across heterogeneous task sizes, we adopt a
dynamic epoch schedule that allocates more training epochs to tasks containing
more classes.  
The number of epochs for task $t$ is determined by
\begin{equation}
E(t)
=
\mathrm{clamp}\!\Bigl(
E_{\min},\ 
E_0 \left(\frac{|\mathcal{Y}_t|}{t_0}\right)^{\beta},\ 
E_{\max}
\Bigr),
\label{eq:flexe}
\end{equation}
where $t_0 = C/T$ is the average number of classes per task under a balanced
CIL scenario, $E_0$ is the reference epoch count, and $\beta$ controls the
growth rate.  
This schedule ensures that larger tasks with richer intra- and inter-class
variability are trained sufficiently, whereas small tasks are trained more
succinctly but compensated by stronger contrastive regularization.

Together, the task-adaptive contrastive weighting and size-dependent epoch
schedule provide complementary mechanisms for stabilizing representation
learning under step imbalance.  
As shown in Table~\ref{tab:flexe_loss}, removing either component leads to a drop in last-step accuracy, indicating that both the geometric
regularization and the size-aware epoch
allocation contribute positively to performance in SI-CIL.
\begin{table}[t]
\centering
\resizebox{0.7\columnwidth}{!}{%
\begin{tabular}{lcc}
\toprule
Method & CIFAR100 & ImageNet-A  \\
\midrule
w/o $\mathcal{L}_{\mathrm{ctr}}$ & 88.34 & 59.38 \\
w/o $E(t)$ & 88.26 & 58.99 \\
One-A & 88.58 & 59.64 \\
\bottomrule
\end{tabular}
}
\caption{Ablation of the contrastive loss $\mathcal{L}_{\mathrm{ctr}}$ and the size-dependent epoch schedule $E(t)$ in One-A under the SI-CIL setting. Results are last-step accuracy (\%) on CIFAR100 and ImageNet-A.
}
\label{tab:flexe_loss}
\end{table}

\subsection{Computational Complexity}
\label{subsec:com_com}
Consider an adapter layer $\Delta W^{(\ell)} \in \mathbb{R}^{d\times b}$ or 
$\mathbb{R}^{b\times d}$, where $b$ is the adapter bottleneck dimension and 
$d$ is the width of the backbone FFN. During fusion, One-A applies a thin SVD 
to the base adapter,  
$\Delta W_b^{(\ell)} = U_b \Sigma_b V_b^\top$, where the SVD rank 
$r \le \min(d,b)$ denotes the number of retained singular directions. 
This is followed by projecting the aligned adapter into the base subspace,
$V_{a\rightarrow b} = \Delta W_a^{(\ell)\top} U_b \Sigma_b^{-1}$, and 
performing directional fusion 
$V_{\mathrm{final}} = V_b + g \odot (V_{\mathrm{fused}} - V_b)$.
Both the thin SVD and the projection–reconstruction steps scale as 
$\mathcal{O}(d\, r^{2})$, since the dominant operations involve multiplying 
$d\times r$ and $r\times b$ matrices with $r$ singular directions.  
The energy-based gating over the $r$ singular values is  
$\mathcal{O}(r)$ and negligible.  
Thus, each layer incurs a computational cost of $\mathcal{O}(d\, r^{2})$, and
merging an adapter with $L$ layers requires $\mathcal{O}(L\, d\, r^{2})$.
Because One-A performs this merge only once at the end of each task, the total 
additional training overhead across $t$ tasks is $\mathcal{O}(t\, L\, d\, r^{2})$, which is negligible relative to the 
billion-scale FLOPs of backbone forward–backward training.
At inference time, One-A uses only the merged adapter $\Delta W_m$. Therefore, 
the inference-time complexity matches that of a single adapter forward pass and 
does not increase with the number of tasks.

\begin{table}[t]
\centering
\resizebox{0.9\columnwidth}{!}{
\begin{tabular}{lccccc}
\toprule
Dataset & LR & Weight Decay & Batch Size & $E_0$  & $r$\\
\midrule
CIFAR100   & 0.01 & $5\times 10^{-4}$ & 128 & 25 & 128 \\
CUB        & 0.07 & 0.0005 & 16 & 10 & 128 \\
ImageNet-A & 0.08 & 0.003 & 64 & 40 & 128 \\
ImageNet-R & 0.03 & 0.005 & 24 & 30 & 64 \\
\bottomrule
\end{tabular}
}
\caption{Dataset-specific hyperparameters used in both main and supplementary experiments.}
\label{tab:dataset_hyper}
\end{table}

\section{More Experiments}
\label{sec:more_exp}
\subsection{Detailed settings}
\label{subsec:detail_setting}
Our implementation builds on the LAMDA-PILOT codebase~\cite{Zhou_2024,zhou2024continual,sun2025pilot}.  
All data preprocessing follows the original framework without modification.  
For baseline methods, we adopt the best hyperparameter settings reported in their papers or official implementations to ensure fair comparison.
All experiments follow the SI-CIL setup described in
Section~\ref{subsec:sicil}.  
Unless otherwise specified, we use the step-imbalance factor
$\gamma=0.01$, and task sizes are generated using Eq.~\ref{eq:im_ratio}.  
The backbone is a ViT-B/16 pretrained on ImageNet-21K.  
Adapters are inserted into all MLP blocks with bottleneck dimension $r$.
For the contrastive loss in Eq.~\ref{eq:contra_loss}, 
we use a cosine-similarity margin of $\tau = 0.07$. 
The task-size–adaptive weighting in Eq.~\ref{eq:contra_loss_final} 
uses $\lambda_{\min}=0.01$, $\lambda_{\max}=0.1$, and $k=2.3979$.
Contrastive loss is computed within each batch using all positive and negative pairs.
For the dynamic epoch schedule in Eq.~\ref{eq:flexe}, we set the growth exponent to $\beta = 0.5$, which provides a mild sub-linear
scaling of epochs with task size and yields stable training behavior across all
datasets.
Dataset-specific hyperparameters, including the learning rate (LR), 
weight decay, batch size, base number of epochs $E_0$, and the bottleneck dimension $r$ in adapters are provided in Table~\ref{tab:dataset_hyper}.  
Unless otherwise stated, the same configuration is applied to both 
the main experiments and all supplementary ablations.
All results in Table~\ref{tab:AA} and Table~\ref{tab:descending}
are averaged over five random seeds.
Each seed determines a distinct class order and task shuffle,
ensuring that results are not tied to a single SI-CIL sequence.
Figures reporting per-task comparisons are produced under one fixed seed so that
all compared methods share the same task order and class order for fair evaluation.

\noindent \textbf{Task-Agnostic Inference Protocol.}
All evaluations are conducted under a task-agnostic setting. During inference,
the model receives only an input image $x$ and predicts over the accumulated
label space $\mathcal{Y}_{1:t}$.
No task identity, task boundary, or task index is provided.
To avoid any implicit task cues, the test samples from all tasks are
randomly shuffled into a unified evaluation stream. Therefore, the model
cannot exploit task ordering or sample locality to infer the task identity.

For methods that maintain multiple task-specific adapters (e.g., EASE~\cite{zhou2024expandable}, CL-LoRA~\cite{he2025cl} and Per-task Adapter in Table~\ref{tab:ablation_fusion}), inference is also performed without task identity.
Specifically, these methods must retrieve all adapters and select the prediction with the highest confidence, following their original design.
This ensures that all methods are compared under the same task-agnostic
protocol.
Our One-A model uses a single merged adapter and directly performs
classification over $\mathcal{Y}_{1:t}$ without any retrieval or routing,
resulting in constant inference cost with respect to the number of tasks.

\noindent \textbf{Additional discussion on evaluation metrics.}
Following the standard evaluation protocol of~\cite{rebuffi2017icarl}, 
we report the last-step accuracy $A_T$ and the average accuracy $\bar{A}$.
The average accuracy is defined as the arithmetic mean of per-step accuracies:
\begin{equation}
\bar{A} = \frac{1}{T} \sum_{t=1}^{T} A_t,
\end{equation}
where $A_t$ denotes the accuracy over all seen classes after learning task $t$.
This metric is widely adopted in class-incremental learning and is appropriate when each task introduces the same number of classes, such that all steps contribute equally.

In addition, we evaluate the degree of catastrophic forgetting using the forgetting metric $F$~\cite{chaudhry2018riemannian}, which quantifies the loss in performance on earlier tasks after the model trained on subsequent tasks.
With $T$ incremental tasks, let $A_{j,k}$ denote the accuracy on task $j$ after the model has been trained up to task $k$ ($k \ge j$). 
The forgetting of task $j$ after learning task $T$ is defined as
\begin{equation}
f_j = \max_{k \in \{j, \dots, T-1\}} A_{j,k} - A_{j,T},
\quad \forall j < T.
\end{equation}
The overall forgetting is computed as the average over all tasks:
\begin{equation}
F = \frac{1}{T-1} \sum_{j=1}^{T-1} f_j.
\end{equation}

\subsection{Results under the Descending Step-Imbalanced Setting}
\label{subsec:descending}

\begin{figure*}[t]
    \centering
    \includegraphics[width=\linewidth]{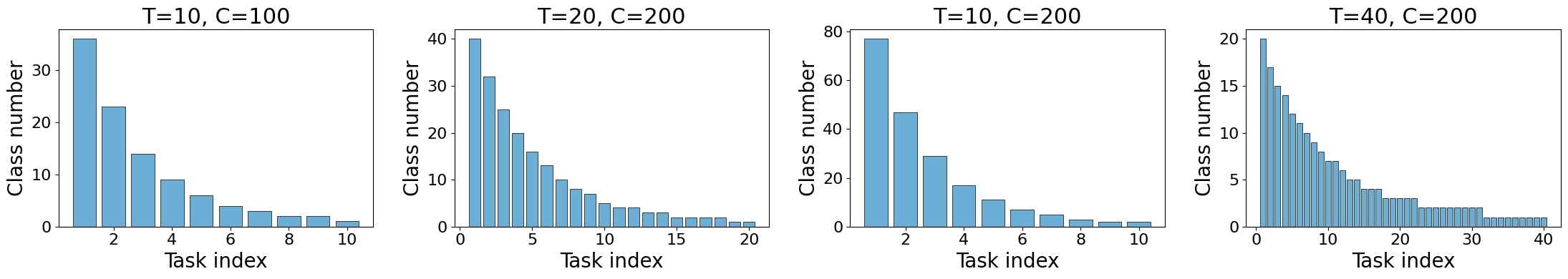}
    \caption{Class-per-task distributions for the descending SI-CIL configurations.}
    \label{fig:descending_distribution}
\end{figure*}

\begin{table*}[t]
\centering
\resizebox{\textwidth}{!}{%
\begin{tabular}{l|cccc}
\Xhline{0.8pt}
\textbf{Method}
& CIFAR100 (T=10)
& CUB (T=20)
& ImageNet-A (T=10)
& ImageNet-R (T=40) \\
\hline
L2P~\cite{wang2022learning}               & 76.16 ± 1.76 & 53.47 ± 2.39 & 36.71 ± 1.29 & 46.43 ± 1.01 \\
DualPrompt~\cite{wang2022dualprompt}      & 71.38 ± 0.54 & 55.19 ± 0.98 & 33.16 ± 2.12 & 46.28 ± 0.60 \\
CODA-Prompt~\cite{smith2023coda}          & 80.01 ± 0.48 & 64.83 ± 0.95 & 45.93 ± 0.37 & 54.73 ± 0.22 \\
Aper-adapter~\cite{zhou2025revisiting}    & 88.24 ± 0.72 & \textcolor{blue}{\textbf{87.67 ± 0.19}} & 53.63 ± 0.60 & 62.79 ± 0.34 \\
EASE~\cite{zhou2024expandable}            & 86.56 ± 1.16 & 84.31 ± 0.45 & \textcolor{blue}{\textbf{56.16 ± 0.28}} & 65.00 ± 0.77 \\
SEMA~\cite{wang2025self}                  & \textcolor{blue}{\textbf{88.31 ± 0.74}} & 67.88 ± 1.97 & 43.95 ± 1.98 & 64.92 ± 1.02 \\
ACMap~\cite{fukuda2025adapter}            & 83.25 ± 0.18 & 86.57 ± 0.41 & 48.96 ± 0.71 & 67.62 ± 0.34 \\
CL-LoRA~\cite{he2025cl}                   & 80.08 ± 1.59 & 49.22 ± 4.39 & 53.79 ± 1.95 & \textcolor{blue}{\textbf{67.85 ± 0.84}} \\
\hline
One-A (Ours) 
& \textcolor{red}{\textbf{89.61 ± 0.06}} {\small(+1.30)}
& \textcolor{red}{\textbf{88.55 ± 0.05}} {\small(+0.88)}
& \textcolor{red}{\textbf{61.27 ± 0.12}} {\small(+5.11)}
& \textcolor{red}{\textbf{73.35 ± 2.44}} {\small(+5.50)} \\
\Xhline{0.8pt}
\end{tabular}
}
\caption{Last-task accuracy ($A_T$) under the descending scenario with step-imbalance factor $\gamma=0.01$. 
All results are averaged over 5 runs with mean $\pm$ standard deviation. \textcolor{red}{\textbf{Red}} denotes the best result and \textcolor{blue}{\textbf{Blue}} denotes the second best in each column.
Numbers in parenthesis indicate the improvement over the second-best method.}
\label{tab:descending}
\end{table*}

In addition to the randomly permuted SI-CIL configuration introduced in
Section~\ref{subsec:sicil}, we also evaluate a \emph{descending} variant in which
tasks are ordered strictly from large to small according to their class
counts.  
This removes the random permutation step used in Figure~\ref{fig:step_imbalance_generation}
and produces a head-to-tail curriculum where the model first
encounters the most diverse tasks and gradually transitions to tasks containing
only a few classes.
This setting is not only a natural complement to the default SI-CIL setup but
also reflects many real-world scenarios where data arrives in a progressively
narrowing form.  
For example, large-scale pretraining or early-stage deployment typically exposes
a model to broad, high-diversity data, while later updates may consist of
narrow or domain-specific classes (e.g., adding a few rare categories to an
existing recognition system).  
Evaluating under the descending curriculum therefore captures a practically
relevant regime where forgetting and subspace distortion tend to be more
pronounced.

Table~\ref{tab:descending} summarizes the last step accuracy $A_{T}$ on four datasets under this
descending configuration as illustrated in Figure~\ref{fig:descending_distribution}.  
Across all settings, One-A achieves consistent improvements over all benchmarks. In particular, it shows an improvement of 5.11\% and 5.50\% on ImageNet-A and ImageNet-R respectively.
These results illustrate the core strength of One-A. Its asymmetric subspace
alignment protects the dominant directions learned from large early tasks from
being overwritten by the noisy or low-rank updates of small later tasks, while
the information-adaptive fusion mechanism appropriately balances the influence
of head and tail tasks throughout training.  
In addition, the directional gating module stabilizes the integration process
by modulating updates along important singular directions, allowing new
information to be incorporated without compromising established structure.

\subsection{Results under the Balanced Setting}
\label{subsec:balance}

\begin{table*}[t]
\centering
\resizebox{\textwidth}{!}{%
\begin{tabular}{l|cc|cc|cc|cc}
\Xhline{0.8pt}
\multirow{2}{*}{\textbf{Method}} 
& \multicolumn{2}{c|}{T=5} 
& \multicolumn{2}{c|}{T=10} 
& \multicolumn{2}{c|}{T=20} 
& \multicolumn{2}{c}{T=40} \\
& $A_{T}$ & $\bar{A}$ 
& $A_{T}$ & $\bar{A}$
& $A_{T}$ & $\bar{A}$ 
& $A_{T}$ & $\bar{A}$ \\
\hline
L2P~\cite{wang2022learning}              & 62.05 ± 1.81 & 74.68 ± 1.62 & 59.22 ± 1.14 & 73.24 ± 0.38 & 57.28 ± 0.87 & 70.04 ± 0.27 & 55.19 ± 0.95 & 66.77 ± 1.03 \\
DualPrompt~\cite{wang2022dualprompt}    & 63.07 ± 0.49 & 75.41 ± 0.75 & 57.51 ± 0.96 & 72.52 ± 1.70 & 65.09 ± 1.12 & 77.32 ± 0.45 & 56.25 ± 1.89 & 71.15 ± 1.80 \\
CODA-Prompt~\cite{smith2023coda}    & 78.50 ± 0.42 & 86.14 ± 0.35 & 74.26 ± 0.15 & 83.30 ± 0.64 & 70.96 ± 1.43 & 81.85 ± 0.29 & 65.60 ± 1.18 & 77.53 ± 0.14 \\
Aper-adapter~\cite{zhou2025revisiting}   & \textcolor{blue}{\textbf{87.67 ± 0.19}} & \textcolor{blue}{\textbf{91.29 ± 0.74}} & \textcolor{red}{\textbf{87.28 ± 0.14}}  & \textcolor{red}{\textbf{91.79 ± 0.63}} & 86.89 ± 0.32 & 91.76 ± 0.57 & 87.19 ± 0.01 & 91.93 ± 0.63 \\
EASE~\cite{zhou2024expandable}          & 84.34 ± 0.33 & 89.51 ± 0.73 & 86.01 ± 0.23  &  90.83 ± 0.61 & \textcolor{blue}{\textbf{87.11 ± 0.13}} & \textcolor{blue}{\textbf{91.79 ± 0.52}} & 86.79 ± 0.39 & 91.88 ± 0.43 \\
SEMA~\cite{wang2025self}          & 80.17 ± 0.17 & 87.43 ± 0.42 & 75.72 ± 0.40 & 84.60 ± 0.08 & 74.32 ± 1.14 & 83.57 ± 0.36 & 64.67 ± 0.39 & 77.42 ± 0.85 \\
ACMap~\cite{fukuda2025adapter} 
              & 87.05 ± 0.35 & 91.06 ± 0.51 & 83.11 ± 0.35 & 90.03 ± 0.69 & 86.92 ± 0.35 & 91.58 ± 0.56 & \textcolor{blue}{\textbf{87.47 ± 0.13}} & \textcolor{red}{\textbf{92.10 ± 0.75}} \\
CL-LoRA~\cite{he2025cl}       & 84.45 ± 0.31 & 89.67 ± 0.53 & 74.32 ± 1.04 & 85.07 ± 1.01 & 66.54 ± 1.72 & 77.94 ± 1.70 & 58.64 ± 1.03 & 69.19 ± 0.89 \\
\hline
One-A (Ours)  & \textcolor{red}{\textbf{88.06 ± 0.35}} & \textcolor{red}{\textbf{91.67 ± 0.48}} & \textcolor{blue}{\textbf{87.14 ± 0.24}} & \textcolor{blue}{\textbf{91.28 ± 0.71}} & \textcolor{red}{\textbf{87.31 ± 0.15}} & \textcolor{red}{\textbf{92.32 ± 0.53}} & \textcolor{red}{\textbf{87.52 ± 0.13}}  & \textcolor{blue}{\textbf{92.01 ± 0.29}}\\
\Xhline{0.8pt}
\end{tabular}
}
\caption{Last-task accuracy ($A_T$) and average accuracy $\bar{A}$ under the balanced scenario with CUB under different task numbers.  
All results are averaged over 5 runs with mean $\pm$ standard deviation. \textcolor{red}{\textbf{Red}} denotes the best result and \textcolor{blue}{\textbf{Blue}} denotes the second best in each column.}
\label{tab:balance_cil_summary}
\end{table*}

\begin{table*}[t]
\centering
\resizebox{0.7\textwidth}{!}{%
\begin{tabular}{l|cc|cc}
\Xhline{0.8pt}
\multirow{2}{*}{\textbf{Method}} 
& \multicolumn{2}{c|}{CIFAR100 (T=10)} 
& \multicolumn{2}{c}{CUB (T=20)} \\
& $A_{T}$ & $\bar{A}$ 
& $A_{T}$ & $\bar{A}$ \\
\hline



Aper-adapter~\cite{zhou2025revisiting} & 78.91 ± 0.63 & 85.24 ± 2.02 & 79.30 ± 0.02 & 85.78 ± 0.59 \\

EASE~\cite{zhou2024expandable} & \textbf{81.20 ± 1.53} & \textbf{88.50 ± 0.95} & 79.40 ± 0.27 & 86.59 ± 0.73 \\

InfLoRA~\cite{liang2024inflora} & 71.69 ± 1.55 & 79.81 ± 2.42 & 49.32 ± 1.93 & 62.52 ± 2.99 \\

SEMA~\cite{wang2025self} & 64.41 ± 0.85 & 74.98 ± 2.15 & 49.06 ± 0.16 & 62.00 ± 1.53 \\

ACMap~\cite{fukuda2025adapter} & 80.50 ± 0.13 & 87.11 ± 0.83 & 79.47 ± 0.04 & 85.90 ± 0.63\\

CL-LoRA~\cite{he2025cl} & 76.17 ± 3.49 & 86.02 ± 2.29 & 67.90 ± 0.97 & 81.41 ± 1.65\\

\hline
One-A (Ours) &81.05 ± 0.40 & 86.65 ± 0.86 & \textbf{81.10 ± 0.42}& \textbf{87.58 ± 1.06}\\
\Xhline{0.8pt}
\end{tabular}
}
\caption{Last task accuracy ($A_{T}$) and average accuracy ($\bar{A}$) under the long-tailed step-imbalance setting.}
\label{tab:lt}
\end{table*}
We also conducted experiments in the balanced setting, where each task introduces an equal number of classes. As CUB is one of the most widely used and representative fine-grained benchmarks in class-incremental learning, we provide a comprehensive evaluation on CUB under four task lengths $T=5$, $T=10$, $T=20$ and $T=40$. This setup allows us to assess not only overall performance, but also the robustness of each method as the task sequence becomes longer.
As shown in Table~\ref{tab:balance_cil_summary}, One-A consistently achieves strong last-task accuracy $A_T$ and average accuracy $\bar{A}$ across all task numbers. Notably, our method maintains top performance even when the number of tasks increases to $T=40$, where the incremental steps become significantly smaller. This demonstrates that One-A scales reliably with task length and remains stable under both short-task and long-task regimes.

These results further validate that One-A not only performs well in step-imbalanced settings, but also generalizes effectively in balanced scenarios. The fact that our method consistently delivers strong performance with low inference cost highlights its robustness and efficiency, making it a promising solution for both step-imbalanced and balanced class-incremental learning tasks.

\begin{figure}[t]
    \centering
    \includegraphics[width=0.7\linewidth]{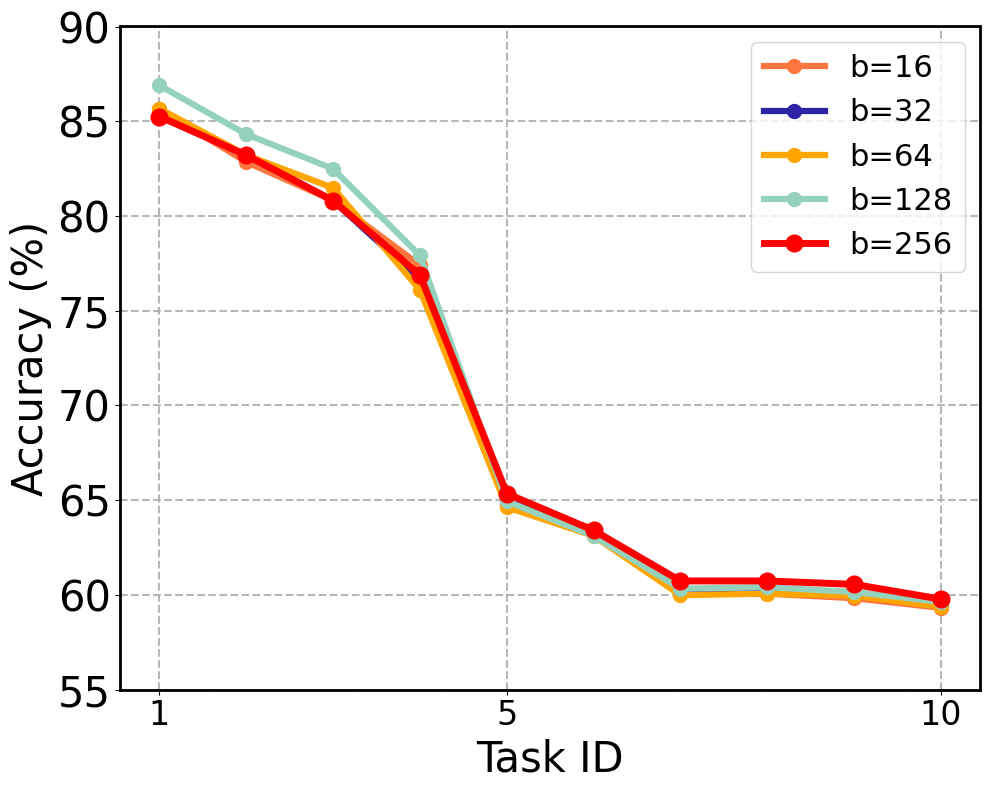}
    \caption{Adapter rank sensitivity on ImageNet-A with \( T = 10 \) tasks.}
    \label{fig:rank_compare}
\end{figure}

\begin{figure}[t]
    \centering
    \includegraphics[width=0.9\linewidth]{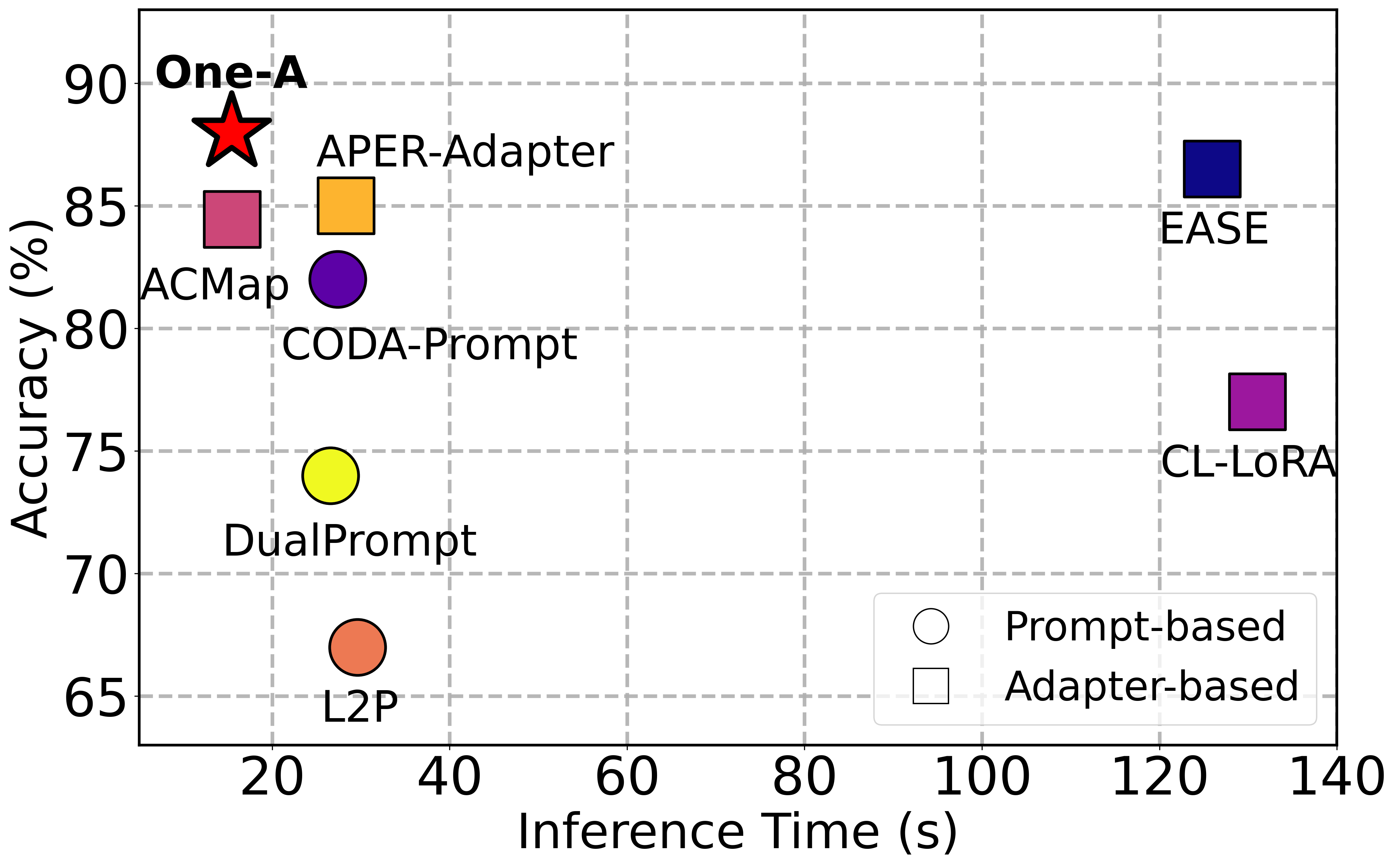}
    \caption{Inference accuracy and efficiency on CIFAR100 under SI-CIL}
    \label{fig:inference_time}
\end{figure}

\subsection{Adapter Rank Sensitivity} 
\label{subsec:adapter_rank}
We study the influence of the adapter bottleneck dimension by testing multiple ranks \( b \in \{16, 32, 64, 128, 256\} \) on ImageNet-A with \( T = 10 \) tasks.
Figure~\ref{fig:rank_compare} shows that all configurations exhibit very similar accuracy along the task sequence. Although the curves slightly cross at some early tasks, the gap among different ranks quickly shrinks, and the final accuracies are nearly the same.
These results indicate that One-A remains stable with respect to the choice of adapter rank.  
The method maintains similar behavior across a wide range of capacities, and the bottleneck dimension does not have a meaningful effect on its performance.

\subsection{Inference Efficiency}
\label{subsec:inference_eff}
A practical benefit of One-A is that inference always uses a single
adapter, regardless of how many tasks have been encountered.  
This eliminates the multi-adapter stacking and
leads to consistently lower latency at test time.
Because only one adapter is active during inference, runtime remains comparable
to a standard single-task adapter and substantially faster than methods whose
inference cost scales with the number of tasks.
Figure~\ref{fig:inference_time} illustrates this efficiency--accuracy trade-off. One-A achieves the best accuracy while occupying the lowest-latency region of
the spectrum, demonstrating its advantage as both an effective and efficient
SI-CIL solution.

\section{Long-Tailed Step-Imbalance: Additional Results and Future Work}
\label{sec:ltsicil}

The main paper focuses on step imbalance, where the number of classes varies across tasks while the sample distribution within each class is balanced.
This setting isolates task-level imbalance to study its effect on subspace merging and incremental adaptation.
However, real-world scenarios may involve additional sample-level imbalance, leading to long-tailed data distributions within each task.

To evaluate robustness under mixed imbalance, we construct a long-tailed step-imbalance setting where both the number of classes per task and the sample distribution within each class are imbalanced.
Table~\ref{tab:lt} reports the results on CIFAR100 and CUB.
Although our method is not specifically designed for long-tailed learning, it remains competitive under this mixed setting and achieves the best performance on CUB.
On CIFAR100, while EASE obtain slightly higher accuracy, our approach still delivers strong results and maintains a substantially lower inference cost due to the use of a single merged adapter, as shown in Figure~\ref{fig:inference_time}. Note that the inference time remains identical to the step-imbalance setting because the test set is class-balanced, highlighting the efficiency advantage of our framework under both settings.
Furthermore, extending our framework to simultaneously handle class-level and sample-level imbalance in continual learning is an important direction for future work.


\end{document}